\pdfoutput=1
\def\isarxiv{1}
\documentclass{article}

\usepackage{microtype}
\usepackage{graphicx}
\usepackage{booktabs}
\usepackage{makecell}
\usepackage{array}
\usepackage{adjustbox}
\usepackage{amsmath}
\usepackage{amssymb}
\usepackage{xspace}
\usepackage{hyperref}
\newcommand{\hexurlbreaks}{\expandafter\def\expandafter\UrlBreaks\expandafter{\UrlBreaks\do\a\do\b\do\c\do\d\do\e\do\f%
  \do\0\do\1\do\2\do\3\do\4\do\5\do\6\do\7\do\8\do\9}}
\usepackage{algorithm}
\usepackage{algorithmic}
\usepackage{float}

\usepackage{pifont}   % rev8: the numbered marks of the pipeline figure's caption

\ifdefined\isarxiv
  \usepackage[accepted]{mlsys2025}
  \makeatletter
  \renewcommand{\Notice@String}{Preprint. Copyright 2026 by the author(s).}
  \renewcommand{\printAffiliationsAndNotice}[1]{{\let\thefootnote\relax\footnotetext{\hspace*{-\footnotesep}%
    \Notice@String}}}   % the byline sits under the names instead (see the author block)
  \makeatother
  \hypersetup{pdfauthor={Avichal Sahai, Nishant Raj, Animesh Srivastava}}
\else
  \usepackage{mlsys2025}
\fi
\hypersetup{pdfsubject={}}   % the style's default names an earlier conference

\newcommand{\code}[1]{\texttt{#1}}
\newcommand{\fmt}{\code{\#\#\#\#~N}\xspace}
\newcommand{\iv}[2]{\makecell[l]{#1\\{[#2]}}}   % a value over its interval
\newcommand{\pmark}[1]{\textcolor[rgb]{0.835,0.369,0}{\ding{#1}}}   % rev8: the pipeline figure's marks (182-184)
\newif\ifarrangement
\arrangementtrue

\mlsystitlerunning{Lost in the bf16 Cast}

\begin{document}

\twocolumn[
\mlsystitle{Lost in the bf16 Cast: Exporting Ternary Language Models\texorpdfstring{\\}{ }Can Revert Most Low-Learning-Rate Code Changes}

\ifdefined\isarxiv
\begin{mlsysauthorlist}
\begin{tabular}{@{}c@{\hspace{2.5em}}c@{\hspace{2.5em}}c@{}}
\textbf{Avichal Sahai} & \textbf{Nishant Raj} & \textbf{Animesh Srivastava} \\
Ofbusiness & Ofbusiness & Ofbusiness \\
{\fontsize{8}{10}\selectfont\href{mailto:avichal.sahai@ofbusiness.in}{\nolinkurl{avichal.sahai@ofbusiness.in}}} &
{\fontsize{8}{10}\selectfont\href{mailto:nishant.raj@ofbusiness.in}{\nolinkurl{nishant.raj@ofbusiness.in}}} &
{\fontsize{8}{10}\selectfont\href{mailto:animesh.srivastava@ofbusiness.in}{\nolinkurl{animesh.srivastava@ofbusiness.in}}}
\end{tabular}
\end{mlsysauthorlist}
\else
\begin{mlsysauthorlist}
\mlsysauthor{Avichal Sahai}{ofb}
\mlsysauthor{Nishant Raj}{ofb}
\mlsysauthor{Animesh Srivastava}{ofb}
\end{mlsysauthorlist}

\mlsysaffiliation{ofb}{Ofbusiness}
\mlsyscorrespondingauthor{Avichal Sahai}{\href{mailto:avichal.sahai@ofbusiness.in}{\nolinkurl{avichal.sahai@ofbusiness.in}}}
\mlsyscorrespondingauthor{Nishant Raj}{\href{mailto:nishant.raj@ofbusiness.in}{\nolinkurl{nishant.raj@ofbusiness.in}}}
\mlsyscorrespondingauthor{Animesh Srivastava}{\href{mailto:animesh.srivastava@ofbusiness.in}{\nolinkurl{animesh.srivastava@ofbusiness.in}}}

\fi

\mlsyskeywords{ternary language models, quantization-aware training, model export, bf16, deployment}

\vskip 0.3in

\begin{abstract}
Ternary language models such as BitNet b1.58, Falcon-E and BitCPM are fine-tuned through higher-precision latents
and deployed as ternary codes by an export that, in the labs' documented pipelines, first casts latents to bf16. We
audit releases from three labs. In released checkpoints, fp32 quantization of the shipped latents disagrees
with deployed codes on 0.83--1.77\% of codes (Falcon-E, BitCPM) and 1.530\% (2B-4T); most Falcon-E and BitCPM
disagreements are bf16 products exactly on the threshold, which ties-to-even maps to zero, and the unmodified
onebitllms exporter reproduces all four Falcon-E releases byte for byte. At fine-tuned endpoints with ULP-matched
rates, the documented export lowers greedy GSM8K strict accuracy from 58.79\% to 0.78\% (Falcon-E-1B-Base) and
36.13\% to 0.39\% (BitCPM-CANN-0.5B), and a bf16 save and reload lowers 2B-4T's by 27.54 points while its
last-number accuracy rises at that endpoint. The strict decrease recurs at each of three new seeds in all three
models and has the same direction on 607 reserve test rows, but it is recipe-dependent: at Falcon-E's
online-loss-selected rate the export was non-inferior, and after a longer recipe no difference was detected. Two
remedies, writing the trained codes directly or adjusting bf16 inputs so unchanged tools emit them, met an inherited
4-point strict-accuracy non-inferiority criterion in all three models, also for the pooled mean over three fixed new
trajectories per model and, separately, on the reserve rows. In a frozen study of bf16 latents in fine-tuning (one task, one
trajectory per arm), native bf16 writes left 2B-4T's endpoint codes unchanged and changed one Falcon-E code;
emulated stochastic-rounding and compensated writers changed codes and were non-inferior to fp32 training at that
margin. In two model families, randomized threshold-distance interventions support distance-dependent selection of
changed codes.
\end{abstract}
]

\printAffiliationsAndNotice{\raggedright}   % the argument is used only in the named build
\chead{\small\bf Lost in the bf16 Cast}   % independent of the style's running-title height test

\section{Introduction}
\label{sec:intro}

Ternary language models store each linear weight as a code in $\{-1,0,+1\}$ with a per-tensor or per-row scale
\citep{ma2024era,ma2025bitnet}. Labs release them in two forms: a latent checkpoint in bf16, from which
quantization-aware fine-tuning starts, and a deployed checkpoint of codes and scales, produced by an export step.
Fine-tuning updates higher-precision latent weights, and a code changes only when its latent crosses a threshold at half
the mean magnitude of its tensor (of its row, for BitCPM). At low learning rates most changed latents finish just past
that threshold (\S\ref{sec:endpoints}). The labs' documented exports pass the latents through bf16, whose 8-bit
significand spaces representable values roughly 0.4--0.8\% apart near the threshold, so the cast can return a
just-crossed latent to its starting value (Figure~\ref{fig:mechanism}). The model evaluated during fine-tuning is then
not the model that ships.

\begin{figure*}[t]
\centering
\includegraphics[width=\textwidth]{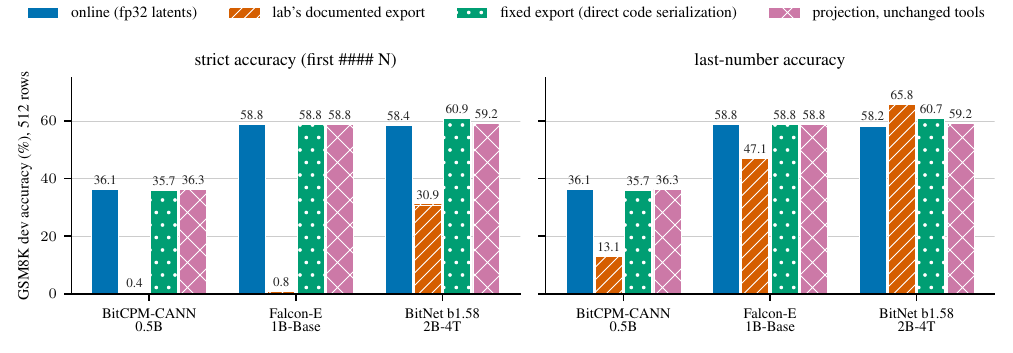}
\caption{GSM8K dev accuracy (512 rows, greedy) of one fine-tuned endpoint per model: evaluated online from the fp32
latents, after the lab's documented export, after the fixed export (direct code serialization) and after the
projection that lets the unchanged tools emit the online codes (ULP-compensated input projection). Strict accuracy takes the first \fmt; the last-number rule
takes the last number. The documented paths differ by model (\S\ref{sec:endpoints}). Bars are descriptive point
estimates; the paired contrasts and their intervals are in Tables~\ref{tab:endpoints} and \ref{tab:remedies}.}
\label{fig:remedies}
\end{figure*}

We audit this boundary in the documented pipelines of three labs: TII's onebitllms exporter for Falcon-E
\citep{falcone,onebitllms}, a bf16 checkpoint reloaded with online quantization for Microsoft's BitNet b1.58 2B-4T
\citep{bitnethf,transformers}, and OpenBMB's \code{qat-convert} for BitCPM-CANN \citep{bitcpm}. Our contributions:
\begin{itemize}
\item \textbf{A release audit} (\S\ref{sec:release}). In matched releases, fp32 quantization of the released latents
disagrees with the deployed codes on 0.83--1.77\% of codes (Falcon-E, BitCPM) and 1.530\% (2B-4T). For Falcon-E-1B-Base and
BitCPM most disagreements are bf16 products that land exactly on the rounding threshold, $|w\cdot r| = \tfrac12$ with
$r$ the inverse scale, which ties-to-even maps to zero, and the unmodified onebitllms exporter reproduces all four
Falcon-E releases byte for byte.
\item \textbf{The export at fine-tuned endpoints} (\S\ref{sec:endpoints}). At learning rates selected to match a
nominal learning-rate-to-bf16-ULP ratio across models, the documented export lowers greedy GSM8K strict accuracy from
58.79\% to 0.78\% (Falcon-E-1B-Base) and from 36.13\% to 0.39\% (BitCPM-CANN-0.5B); a bf16 save and reload lowers
2B-4T's strict accuracy by 27.54 points while its last-number accuracy rises by 7.62. The strict decrease recurs at
each of three new seeds in all three models and has the same direction on 607 reserve GSM8K test rows, but it is
recipe-dependent: at Falcon-E's online-loss-selected rate the export was non-inferior and detectably above online,
after a longer Falcon-E recipe no difference was detected, and last-number results differ from strict ones
(\S\ref{sec:newevidence}).
\item \textbf{Two remedies} (\S\ref{sec:remedies}). Writing the training quantizer's captured codes directly into the
deployed representation (direct code serialization), or adjusting the bf16 inputs in bounded steps until the unchanged
tools emit those codes (ULP-compensated input projection), was non-inferior to online evaluation at an inherited
4-point strict-accuracy margin in all three models (Figure~\ref{fig:remedies}), again for the pooled mean over three
fixed new trajectories per model, and separately on the 607 reserve rows.
\item \textbf{bf16 latents during fine-tuning} (\S\ref{sec:training}). In a prospectively frozen study, the on-grid bf16
master-weight study, native bf16
writes left 2B-4T's evaluation-context codes at the endpoint equal to those at step 0, while its strict accuracy rose
55.86 points over the base (last-number $+4.88$ [$-0.39$, $+10.16$]), and changed one Falcon-E code, whose online
last-number accuracy rose 2.54 points while its exported model stayed at 0.20\% strict. Emulated
stochastic-rounding and compensated writers changed codes and were non-inferior to fp32 training at the same 4-point
margin in both models.
\item \textbf{Which codes fine-tuning changes} (\S\ref{sec:selection}). On 2B-4T, changed codes are enriched for
positive first-order base-gradient scores relative to codes at the same stored level, increasingly with initial
distance from the threshold, and randomized interventions that change only the starting distance support
distance-dependent selection in two model families.
\end{itemize}
We claim neither a new rounding mechanism---the cancellation of small updates by bf16 rounding is known, and so are
stochastic and compensated rounding as remedies \citep{gupta2015limited,zamirai2020bf16,miahi2026pulse}---nor a
general export or training penalty. The results concern one recipe family, one to five
trajectories per arm, 512 GSM8K dev prompts, 607 reserve test rows for the headline endpoints and, for three
model/task cases, ARC and BoolQ validation rows. For the prospectively packaged experiments, the analysis and
decision rule were frozen before the run; Appendix~\ref{app:protocol} describes the protocol and lists its exceptions,
including the analyses that were exploratory, adaptive or post hoc.\footnote{Experiments are named descriptively;
Table~\ref{tab:key} (Appendix~\ref{app:protocol}) maps each name and label to the internal registry and its evidence.}

\section{Background}
\label{sec:background}

\textbf{Ternary quantization.} BitNet-style absmean quantization maps a latent tensor $W$ to codes
$q = \mathrm{round}(\mathrm{clamp}(W r,-1,1))$, with amplitude $a = \mathrm{mean}|W|$ and inverse scale $r = 1/a$, per
tensor for Falcon-E and 2B-4T and per row for BitCPM, and deploys $q\cdot a$. A weight's code is nonzero when $|w|$
exceeds $a/2$, that is, when $|w\cdot r| > \tfrac12$. This notation is idealized: the exporters' operation order and
bf16 roundings are part of what \S\ref{sec:release} measures. Quantization-aware training uses a straight-through
estimator: the forward pass uses the codes, and the backward pass updates the latents, kept here in fp32 except in
\S\ref{sec:training}.

\begin{figure}[t]
\centering
\includegraphics[width=\columnwidth]{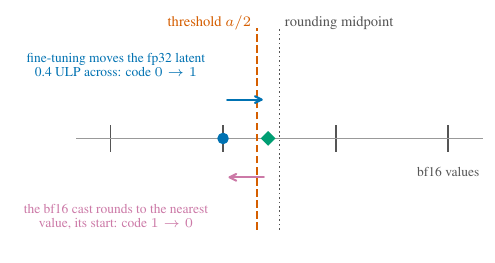}
\caption{Schematic of the reversion at a fixed threshold. The latent starts on a bf16 value 0.3 ULP below the threshold,
and fp32 fine-tuning moves it 0.4 ULP, to 0.1 ULP above. That endpoint is still below the rounding midpoint, inside
its starting value's rounding cell, so the export's round-to-nearest bf16 cast returns it to its start and the code
reverts. An endpoint beyond the midpoint would round onward and keep its new code.}
\label{fig:mechanism}
\end{figure}

\textbf{bf16 near the threshold.} bf16 has an 8-bit significand \citep{kalamkar2019bfloat16}. Near a threshold $t$ the
spacing of representable values, one ULP, is $2^{\lfloor\log_2 t\rfloor-7}$, between 0.4\% and 0.8\% of $t$. Hold the
threshold $t$ fixed, between adjacent bf16 values $L < U$. A code addition that starts at $L$ reverts under
round-to-nearest if its endpoint lies above $t$ but below the rounding midpoint $(L+U)/2$. More generally, a latent
returns to its starting bf16 value whenever its total displacement keeps it inside that value's rounding cell;
overshoot beyond the threshold does not by itself decide this, since an endpoint just past $t$ can lie beyond the
midpoint and round onward. Exact midpoint ties go to the even neighbour (Figure~\ref{fig:mechanism}). The export's
actual threshold also moves with its recomputed scale, which is part of the measured pipeline. Products computed in
bf16 can also land exactly on $|w\cdot r| = \tfrac12$, which \code{torch.round} maps to 0 because it rounds ties to
even.

\begin{figure*}[t]
\centering
\includegraphics[width=\textwidth]{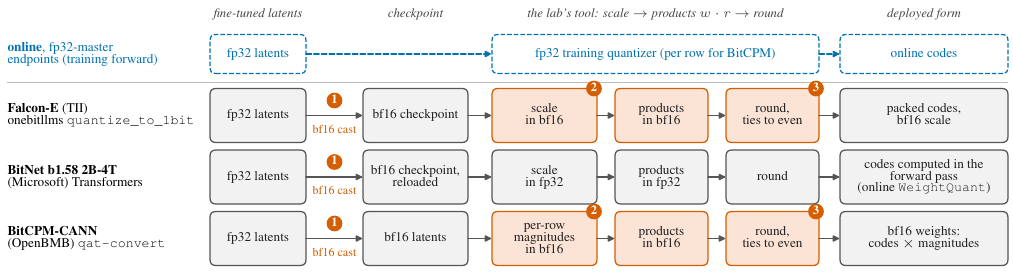}
\caption{The labs' documented export paths from the fp32-master endpoints (schematic) and where bf16 can change a
ternary code. Dashed: those endpoints' fp32 training quantizer, whose online codes both remedies reproduce
(\S\ref{sec:remedies}). \pmark{182}~The bf16 cast can
return a just-crossed latent to its starting value and code (\S\ref{sec:background}, \S\ref{sec:endpoints}).
\pmark{183}~A scale computed in bf16 moves the threshold (\S\ref{sec:release}). \pmark{184}~bf16 products that land
exactly on $|w\cdot r| = \tfrac12$ round to zero under ties-to-even (\S\ref{sec:release}).}
\label{fig:pipeline}
\end{figure*}

\textbf{The documented export paths} (Figure~\ref{fig:pipeline}). All three pass the latents through bf16; two also
compute the scale and the products in bf16.

\section{Released checkpoints encode a different model}
\label{sec:release}

For each matched release we quantized the released bf16 latents with fp32 absmean arithmetic, which is what the
training quantizers of onebitllms and Transformers' online BitLinear compute, and compared the codes with the deployed
ones (Table~\ref{tab:census}). We found no matched latent checkpoint for the reviewed TriLM releases, so they are
excluded; the three families are the matched pairs this search identified, not a census of all releases.

\begin{table}[t]
\centering\scriptsize
\setlength{\tabcolsep}{2.5pt}
\caption{fp32 codes from the released latents against the deployed codes. For Falcon-E and BitCPM the release's own
arithmetic reproduces the deployed codes exactly; for 2B-4T no bf16 recipe we tried does, and its historical source
is not observed.}
\label{tab:census}
\begin{adjustbox}{max width=\columnwidth}
\begin{tabular}{lrrl}
\toprule
release & codes & disagreements & direction \\
\midrule
Falcon-E-1B-Base & 1,585,446,912 & 28,132,333 (1.774\%) & all $\pm1\to0$ \\
Falcon-E-1B-Instruct & 1,585,446,912 & 26,397,811 (1.665\%) & all $\pm1\to0$ \\
Falcon-E-3B-Base & 2,919,235,584 & 31,511,209 (1.079\%) & all $\pm1\to0$ \\
Falcon-E-3B-Instruct & 2,919,235,584 & 28,130,481 (0.964\%) & all $\pm1\to0$ \\
BitCPM-CANN-0.5B & 358,612,992 & 5,996,444 (1.672\%) & all $\pm1\to0$ \\
BitCPM-CANN-3B & 3,229,614,080 & 26,828,322 (0.831\%) & all $\pm1\to0$ \\
BitNet b1.58 2B-4T & 2,084,044,800 & 31,895,216 (1.530\%) & both ways \\
\bottomrule
\end{tabular}
\end{adjustbox}
\end{table}

\textbf{One numerical cause in two labs' export code.} For BitCPM, 95.71\% (0.5B) and 93.87\% (3B) of the
disagreements are weights whose bf16 product has $|w\cdot r|$ exactly 0.5, which \code{torch.round} maps to zero; the
rest have bf16 products just below 0.5, where the bf16 scale moved the threshold. For Falcon-E-1B-Base under
onebitllms' export arithmetic the share is 93.52\%. Every Falcon-E and BitCPM disagreement removes a $\pm1$ code at most
about 1.5 bf16 ULP above the threshold.

\textbf{The actual exporter.} The unmodified onebitllms 0.0.4 \code{quantize\_to\_1bit}, run on each Falcon-E release's
prequantized checkpoint, reproduces all four releases byte for byte in every tensor, including packed codes, weight
scales and non-projection tensors. This is computational provenance, not a recovered export log.

\textbf{2B-4T's hidden codes.} 2B-4T's 31,895,216 disagreements (16,170,283 removed, 15,724,933 added) lie within 0.5
ULP of the threshold in both directions, compatible with rounding of the released latent representation; its original
master and export history are unavailable \citep{bitnet632,ternarycheck2026}. Appendix~\ref{app:release} gives their
bearing on fine-tuning and the C4 loss consequence of the deployed codes, which depends on the release: against fp32
quantization of the released latents, the deployed codes and scales raise cross-entropy for both Falcon-E Base
releases, lower it for 2B-4T and show no detectable difference for the two Instruct releases, so we infer no rule
across releases.

\section{The export at fine-tuned endpoints}
\label{sec:endpoints}

\textbf{Setup.} Each model is fully fine-tuned on GSM8K \citep{cobbe2021gsm8k} for 300 steps with gradient
accumulation 16 (4,800 rows from a 6,833-row training pool), fp32 latents, PagedAdamW8bit, a 30-step warm-up and a
cosine schedule. 2B-4T trains at learning rate $10^{-4}$; Falcon-E-1B-Base and BitCPM-CANN-0.5B train at
$3.125\times10^{-6}$, chosen to match 2B-4T's nominal median ratio of learning rate to bf16 ULP. Each state answers the
same 512 dev prompts, carved from the official training split, by greedy zero-shot decoding. Strict accuracy takes the
first \fmt; the last-number rule takes the last number in the answer. Intervals are 95\% paired row bootstraps (2,000
draws). Each comparison was declared before its run, with one or two trajectories per arm. The nominal ratio is a
rule for choosing the rates; it is not measured optimizer displacement or matched training dynamics. The lower rate was chosen to study threshold-distance effects
at a small nominal learning-rate-to-bf16-ULP ratio, using 2B-4T's nominal median ratio as the reference; it is not a
recommended or typical fine-tuning rate, and Falcon-E's documented rate, $5\times10^{-4}$, is examined below. The Falcon-E
fp32-master runs train through Transformers' online AutoBitLinear, whereas TII's documented training implementation
uses onebitllms' Triton BitNetLinear: the documented exporter is exercised exactly, but the lab's training stack is
not replicated in full.

\begin{table*}[t]
\centering\scriptsize
\setlength{\tabcolsep}{3pt}
\caption{The documented export at fine-tuned endpoints, against online evaluation: strict and last-number accuracy in
percentage points and dev cross-entropy in nats per token (95\% paired row bootstrap). The documented paths are the
onebitllms export (Falcon-E), a bf16 save and reload (2B-4T) and \code{qat-convert} (BitCPM). The last column is the
share of fine-tuning's code changes that the bf16 cast alone returns to the start code (Falcon-E: of code additions; 2B-4T: not measured).}
\label{tab:endpoints}
\begin{adjustbox}{max width=\textwidth}
\begin{tabular}{llrrlllr}
\toprule
model, arm & lr & online & exported & $\Delta$ strict & $\Delta$ last-number & $\Delta$ CE & cast returns \\
\midrule
Falcon-E-1B-Base, A1 & $3.125\times10^{-6}$ & 58.79\% & 0.78\% & \iv{$-58.01$}{$-62.30$, $-53.91$} & \iv{$-11.72$}{$-16.99$, $-6.64$} & \iv{$+0.41513$}{$+0.40251$, $+0.42914$} & 96.3\% \\
Falcon-E-1B-Base, A3 & $5\times10^{-4}$ & 15.82\% & 16.21\% & \iv{$+0.39$}{$-1.56$, $+2.34$} & \iv{$+0.20$}{$-1.76$, $+2.15$} & \iv{$+0.00011$}{$-0.00041$, $+0.00065$} & 0.6\% \\
BitNet 2B-4T, C1 & $10^{-4}$ & 58.40\% & 30.86\% & \iv{$-27.54$}{$-33.01$, $-22.07$} & \iv{$+7.62$}{$+2.54$, $+12.11$} & \iv{$+0.30198$}{$+0.28990$, $+0.31520$} & -- \\
BitCPM-CANN-0.5B, B2 & $3.125\times10^{-6}$ & 36.13\% & 0.39\% & \iv{$-35.74$}{$-40.04$, $-31.84$} & \iv{$-23.05$}{$-27.54$, $-18.95$} & \iv{$+0.21283$}{$+0.20486$, $+0.22082$} & 95.3\% \\
BitCPM-CANN-0.5B, B3 & $10^{-4}$ & 26.37\% & 26.56\% & \iv{$+0.20$}{$-2.93$, $+3.12$} & \iv{$+0.20$}{$-2.93$, $+3.12$} & \iv{$+0.00305$}{$+0.00202$, $+0.00399$} & 4.0\% \\
\bottomrule
\end{tabular}
\end{adjustbox}
\end{table*}

\begin{figure*}[t]
\centering
\includegraphics[width=\textwidth]{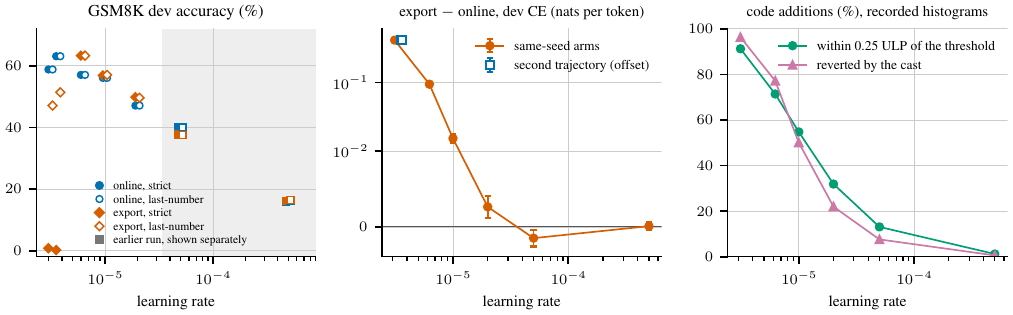}
\caption{Falcon-E-1B-Base, the documented export across fine-tuning learning rates (fp32 latents). Each point is one
trajectory; the second trajectory at the lowest rate is offset to the right. Left: greedy GSM8K dev strict (filled)
and last-number (open) accuracy, online (circles) and exported (diamonds); squares on the shaded band: two arms of an
earlier run, shown separately. Rate selection and this evaluation use the same 512 development rows. Middle: export
minus online dev cross-entropy with 95\% paired row-bootstrap intervals, on a symmetric log scale. Right: descriptive
summaries of each endpoint's recorded margin histogram, the share of code additions within 0.25 bf16 ULP of the
threshold and the share the cast reverts; they are counts, not measurements of latent displacement. Middle and right:
lines connect the declared rates for reading; no curve or between-rate difference was tested.}
\label{fig:rate}
\end{figure*}

\textbf{Falcon-E, through the onebitllms export.} At the low-rate endpoint the documented export raises dev cross-entropy by
$+0.41513$ nats per token, and along the declared export path the projection-latent cast carries the largest increment
($+0.38053$ [$+0.36875$, $+0.39357$]). Of that endpoint's 4,055,607 code additions, 91.2\% lie within 0.25 bf16 ULP of
the threshold, and the cast alone reverts 96.3\% of them (descriptive summaries of the recorded histograms). Greedy
strict accuracy falls from 58.79\% to 0.78\%,
alongside a collapse of adherence to the fine-tuned answer format (97.1\% to 21.7\%); under the last-number rule
accuracy also falls, from 58.79\% to 47.07\%. This is not an erasure of all fine-tuning: under the last-number rule the
exported model stays above the base model by $+23.24$ [$+17.97$, $+28.52$] points. Over the same-seed rates from $3.125\times10^{-6}$ through $5\times10^{-5}$
the cross-entropy cost's point estimates decrease (Figure~\ref{fig:rate}; Appendix Table~\ref{tab:app7}), together with the share of
additions that sit near the threshold, and at $5\times10^{-5}$ the export slightly lowers loss ($-0.00147$ [$-0.00256$, $-0.00034$]). At $5\times10^{-4}$, the documented learning
rate, the contrast is near zero ($+0.00011$ [$-0.00041$, $+0.00065$]), so the full sequence is not monotonic, and
neither its fp32-master arm nor its bf16-parameter arm shows a detectable accuracy difference; the bf16-parameter arm
matches the documented parameter precision within this recipe, not TII's training pipeline. No between-rate
difference or curve was tested. Figure~\ref{fig:rate} (left) shows the accuracies at these rates, and
\S\ref{sec:newevidence} their contrasts. A factorial of the export's two code-change groups at this endpoint is in
Appendix~\ref{app:endpoints}.

\textbf{BitNet 2B-4T, through a bf16 save and reload.} Saving the fine-tuned model as a bf16 checkpoint and reloading it with online
quantization lowers strict accuracy by 27.54 points (C1) and 23.63 points (a second trajectory, C2), while last-number
accuracy rises by 7.62 points in each; adherence to the answer format falls from 98.4\% to 41.4\% in C1. At this original endpoint, strict accuracy and
format adherence fell while last-number accuracy rose; these scoring rules do not isolate formatting changes from
answer-value or reasoning changes.

\textbf{BitCPM, through OpenBMB's \code{qat-convert}.} Served through OpenBMB's own pipeline, the low-rate fine-tune loses accuracy
under both rules. Of its 1,122,949 changed codes the bf16 cast returns 95.3\% to the start code, and the served model
differs from the published checkpoint in 99,080 codes; at a higher rate, see Appendix~\ref{app:endpoints}.

\textbf{Other tasks (Part~T) and an export-arithmetic training forward (Part~DA).} In three model/task recipes
evaluated by fixed prompt-only rank choice with a float32 output head, the declared documented-path comparisons
lowered BoolQ accuracy for Falcon-E by 13.48 points [9.38, 17.58] and for 2B-4T by 4.30 points [1.37, 7.42]. The
Falcon-E ARC validation-mixture contrast was unresolved (documented minus online $-2.93$ [$-6.25$, $+0.39$]); this is
not equivalence or a tested between-task interaction. Both remedies met the inherited four-point non-inferiority
criterion in all six task contrasts, a margin not independently validated for these tasks. Separately, at one
Falcon-E GSM8K recipe, training with the export-arithmetic forward produced a real-tool export non-inferior to the
freshly evaluated ordinary online comparator: strict and last-number differences were both $+2.93$ [$-1.17$, $+7.03$].
Its captured projection codes and scales matched the saved tool output exactly; non-projection casts and inference
arithmetic still differed. Part~T's Falcon-E documented comparator uses the in-process export-arithmetic scaffold; it
was not built by running the real exporter on the unprojected endpoint. Part~T's projected remedy separately uses the
projection's declared tool route. Part~DA constructs both of its documented comparators from the real exporter's
outputs. 2B-4T's documented path is the bf16 save and user reload, and its BoolQ arm was trained on the answer token
only, Falcon-E's arms on the answer and EOS. Each case is one trajectory; ARC \citep{clark2018arc} is the declared Easy and Challenge
validation mixture, not a test score, and validation cross-entropy was recorded during training without selecting
anything, so the rosters were not unseen throughout training. The float32 head is a common measurement convention,
not each model's native bf16 head arithmetic; exact ties count as incorrect, and the intervals are conditional on this
implementation. $\delta = 4$ pp is inherited from the GSM8K dev export contrasts (median 95\% half-width, rounded
down); it was not derived or validated for ARC or BoolQ rank-choice accuracy. Six non-inferior results are not zero
degradation, identical answers, joint coverage or universal remedy efficacy. Part~DA compares two complete training
procedures, at one rate and one trajectory each; it is not an isolated effect of code movement nor a verdict on
deployment-aligned QAT in general, and its non-inferiority is not equivalence or superiority. $\delta = 4$ pp is the
inherited resolution-based margin from the 512-row dev export contrasts (median 95\% half-width, rounded down).
Against the real-tool export of ordinary training, the real-tool export of the model trained with the
export-arithmetic forward scored higher (strict $+60.94$ [$+56.64$, $+65.04$]; last-number $+14.65$ [$+9.77$,
$+19.73$]); this does not show whether it matches ordinary online quality. All intervals are marginal per contrast;
there is no joint coverage claim and no package-level success claim (Appendix Tables~\ref{tab:e35}
and~\ref{tab:e35states}).

\textbf{What these show.} At the low-rate GSM8K endpoints above, the documented exports undo most of fine-tuning's code changes and
can collapse strict-format accuracy, while last-number accuracy falls less or, for 2B-4T, rises; the size of the
effect depends on the learning rate, the export path, the recipe and the scoring rule (\S\ref{sec:newevidence}).

\subsection{New seeds, other recipes and reserve rows}
\label{sec:newevidence}

At the original low-rate recipes, the documented deployment path reduced strict GSM8K accuracy at each of three new
seeds in all three models, and the recorded held-out comparisons had the same strict direction. This effect was
recipe-dependent: Falcon-E's online-CE-selected rate was export-noninferior, while its longer training recipe showed no
detectable export difference. Last-number results differed from strict accuracy, including higher accuracy at two new
2B-4T seeds and an unresolved held-out difference. The two remedies met the declared 4 pp non-inferiority criteria for
the tested comparisons. That criterion is the inherited one defined in \S\ref{sec:remedies}. The held-out rows are 607 seeded-random rows from the 1,119 official GSM8K test indices
remaining after the exposed first 200 (the other 512 of those were exposed earlier); held out from the programme's
training, development and selection, not necessarily from the base models' pretraining; zero-shot prompt, so absolute
accuracies are not comparable with published GSM8K results. These results come from one pre-registered study in three
parts; Appendix Tables~\ref{tab:e34} and \ref{tab:e34rem} list all 23 of its primary contrasts, every one evaluable.

\textbf{Falcon-E rates (Part A).} On a second trajectory at A1's rate (a different seed and 4,800-row draw), the
documented export lowered strict accuracy from 63.09\% to 0.20\% (last-number from 63.09\% to 51.37\%; $\Delta$ strict
$-62.89$ [$-67.19$, $-58.59$]; last-number $-11.72$ [$-16.80$, $-6.64$]); the strict-accuracy loss is accompanied by a
last-number loss. At $6.25\times10^{-6}$, the rate with the lowest online dev loss among those tested (one trajectory
sharing A1's seed, draw and order; development rows also used for the selection), the documented export raised strict
accuracy, non-inferior and detectably above online at $\delta = 4$ pp ($\Delta$ strict $+6.25$ [$+1.76$, $+10.94$];
last-number $+6.25$ [$+1.76$, $+10.94$]), so an unqualified claim that this export necessarily degrades fine-tuned task
accuracy is false. $\delta = 4$ pp is the inherited resolution-based margin derived from these 512 dev rows' export
contrasts (median 95\% half-width, rounded down); it is used for continuity, not re-derived. The documented export at
$2\times10^{-5}$, the rate with the lowest exported dev loss among those tested, was detectably below the
$3.125\times10^{-6}$ online model by more than 4 points (different models; development rows also used for the
selection; $\Delta$ strict $-8.98$ [$-13.67$, $-4.49$]; last-number $-9.18$ [$-13.87$, $-4.69$]; the strict-accuracy
loss is accompanied by a last-number loss). This does not establish that no deployed rate can match online accuracy;
the descriptive best-deployed-minus-best-online readout over the five in-run endpoints is $+0.20$ points,
over different selected endpoints, with no interval and no label, and neither readout validates a tuning procedure on
independent data. Part A's intervals describe these fixed endpoints on the reused development rows. Choosing one rate
per question avoids a later choice among accuracy results; it does not make the rows independent of the CE-based
selection, it supplies no selection-adjusted coverage, and it is not a validation of a tuning procedure.

\textbf{New seeds and other recipes (Part B).} At each of three new training seeds (each a new 4,800-row draw and
order), each model's documented path lowered strict accuracy, so the strict decrease recurs at each of three new seeds
in all three models (every strict interval below zero; per-seed strict and last-number contrasts in Appendix
Table~\ref{tab:e34}). For Falcon-E and BitCPM the
strict-accuracy loss is accompanied by a last-number loss at every seed; for 2B-4T, last-number accuracy was higher at
two seeds ($+8.98$ [$+4.30$, $+13.67$] and $+7.42$ [$+2.54$, $+12.50$]) and unresolved at the third ($+4.49$
[$-0.20$, $+9.18$]). At these seeds, 2B-4T's documented-path generations stopped at EOS on 8.6--16.0\% of rows
(online 98.0--98.8\%) and reached the token cap on 84.0--91.4\%; these rates describe the generations, and no causal
allocation between format and ability is made. Appendix Tables~\ref{tab:five} and \ref{tab:e34fmt} list each model's five trajectories, not a
seed-variance estimate, and the format, stop and cap rates. After 1,200 steps at A1's
rate (A-long; about 2.8 passes over the pool; one trajectory), no difference in strict accuracy was detected after the
documented export ($\Delta$ strict $+1.95$ [$-2.34$, $+6.25$]; last-number $+2.15$ [$-2.15$, $+6.45$]); no detectable
difference is not equivalence, and not evidence that the export is harmless for task use, on these 512 dev prompts, in
this recipe. A-long changes steps, data exposure and the cosine schedule together; it answers whether the export effect
occurs after this longer recipe, not the effect of duration alone. For 2B-4T at $3\times10^{-5}$, $10^{-4}$ and
$3\times10^{-4}$ (A-BN; one trajectory each; simultaneous intervals over the three strict contrasts), the bf16 save
lowered strict accuracy at the first two, including the pre-registered $10^{-4}$, which had the lowest online dev loss
of the three, and raised it at the third ($\Delta$ strict/last-number $-64.84$/$-5.08$, $-27.54$/$+7.62$ and
$+8.98$/$+9.18$ points; Appendix Table~\ref{tab:e34}).

\textbf{Reserve rows (Part C).} On the 607 reserve rows the documented path lowered strict accuracy in all three models:
Falcon-E from 53.87\% to 0.33\% ($\Delta$ strict $-53.54$ [$-57.50$, $-49.59$]; last-number $-12.52$ [$-17.46$,
$-7.58$]), 2B-4T from 52.88\% to 25.37\% ($-27.51$ [$-32.29$, $-22.73$]; last-number $+3.95$ [$-0.33$, $+8.24$],
unresolved) and BitCPM from 31.96\% to 0.16\% ($-31.80$ [$-35.58$, $-28.01$]; last-number $-19.77$ [$-24.05$,
$-15.49$]). For Falcon-E and BitCPM the strict-accuracy loss is accompanied by a last-number loss. These contrasts do not
show whether reasoning was kept or lost, nor that the strict loss is purely a format effect.

\textbf{Remedies (Parts B and C).} All six pooled remedy records of Part B and all six remedy records of Part C were
non-inferior to online evaluation at $\delta = 4$ pp. $\delta = 4$ pp is the inherited resolution-based margin from
the 512-row dev export contrasts (median 95\% half-width, rounded down); it was not re-derived for the 607 reserve
rows. 2B-4T's pooled fixed export was non-inferior and detectably above online, with its lower bound, $+0.20$, at the
boundary flag below; each pool's three per-seed contrasts are secondary (Appendix Table~\ref{tab:e34rem}). The pooled estimand averages each
row across the three fixed trajectories and resamples rows together; it answers a finite three-trajectory question,
not uncertainty over future seeds. By construction, the fixed export reproduces the online codes, so direct code
serialization measures the residual left by deployed-inference arithmetic and bf16 non-projection parameters (and,
for Falcon-E, the deployed format's reciprocal scale); its positive outcome is expected by construction. By
construction, once the codes match, the projected model carries the online codes; the remaining difference is the
exporter's own bf16 scale and the deployed arithmetic. As in \S\ref{sec:remedies}, the compared paths also cast the
non-projection parameters; the scale representation, inference arithmetic and these casts are not separated, and
matching codes is not matching the function or the generated answers. Non-inferiority is not equivalence, identical generations, zero
degradation or unconditional restoration. Four primary records carry the declared boundary flag, an interval endpoint within
0.4 pp of a threshold: the pooled BitCPM projection [$-0.20$, $+1.63$], the pooled 2B-4T fixed export [$+0.20$,
$+2.41$] and projection [$0.00$, $+2.35$], and 2B-4T's fixed export on the reserve rows [$-3.13$, $+0.33$]. These
flags arise at zero, the sign threshold, not at the $-4$ pp non-inferiority boundary; exact contact with zero remains
no detected difference under the strict-inequality sign rule. Intervals are marginal per contrast except A-BN's
declared three-contrast simultaneous strict family. Last-number intervals are outside that family. No joint coverage
across the package or package-level success is claimed.

\subsection{Which codes fine-tuning changes}
\label{sec:selection}

\textbf{Why low-rate changes sit at the threshold.} A code changes only when its latent crosses the threshold, and a
latent that starts near it needs only a small displacement to cross. The low-rate endpoints show crossings of this
kind: on 2B-4T fine-tuned on GSM8K, 12,905,594 of 2,084,044,800 codes change, 99.97\% of them at the stored level
nearest the threshold, and at Falcon-E's low-rate endpoint 91.2\% of the code additions lie within 0.25 bf16 ULP of it
(Figure~\ref{fig:rate}, right). These counts describe concentration near the threshold. At the measured Falcon-E
endpoint, the cast separately reverted 96.3\% of additions. Reversion depends on the latent's rounding cell and the
quantizer's scale, not threshold proximity alone.

\textbf{Which crossings the base gradient favours.} Each eligible coordinate, within 2 bf16 ULP of the threshold, gets
the base model's first-order prediction of the loss reduction from its code change, $b_i = -g_i\,a\,\Delta q_i$.
Same-level lift is the share of positive scores among crossers, the coordinates whose code changed, minus that share
among all eligible coordinates of the same tensor, class and stored magnitude (Appendix~\ref{app:selection}). Lift
rises with initial distance (Appendix Figure~\ref{fig:lift}): from $+0.063$ below 0.1 ULP to $+0.157$ at 0.5--1 ULP. The
distance contrast, the lift at 0.5--1 ULP minus the lift below 0.1 ULP, is $+0.094$ [$+0.081$, $+0.111$] on GSM8K, in a
95\% tensor-cluster bootstrap conditional on the fixed trajectory, and positive at a second seed and, under rules
declared before their outcomes, on BoolQ \citep{clark2019boolq} and MATH \citep{hendrycks2021math}
(Appendix Table~\ref{tab:app12}).

\textbf{Farther crossings.} Randomized interventions that move matched pairs of coordinates nearer to or farther from
the threshold in the fp32 latent, with the starting computation bitwise identical, raise the far-moved members' lift
over the near-moved members' by $+0.0372$ and $+0.0391$ at two 2B-4T seeds and by $+0.0885$ on Falcon-E-1B-Base,
below its frozen prediction (Appendix Table~\ref{tab:intervention}). They support distance-dependent selection: far-moved
crossings are more enriched for positive scores, relative to their level-mates, than near-moved ones. Far placement also makes
crossing less likely, so this concerns what the remaining crossings are enriched for, not more useful edits; support
rests on an outcome-dependent rule that excludes about 34\% of the pair weight; and no benefit for individual
coordinates, new generic threshold mechanism or improved task performance is established
(Appendix~\ref{app:selection}).\ifarrangement{} A related study that rearranges Falcon-E's released latents is in
Appendix~\ref{app:arrangement}.\fi

\section{Two remedies}
\label{sec:remedies}

The original remedy studies start from one existing fine-tuned endpoint per model and its online codes---BitCPM-CANN-0.5B's B2, and
Falcon-E's A1 and 2B-4T's C1 retrained to their pinned endpoints with the fp32 latents retained---and compare a deployed
state with the online state on the same 512 dev rows.

\textbf{Direct code serialization} (the fixed export) writes the codes captured under the fp32 training quantizer into each model's
declared deployed representation and evaluates it in bf16 without autocast. It replaces the export step.

\textbf{ULP-compensated input projection} (the projection) keeps each lab's tool unchanged and adjusts its input
(Appendix~\ref{app:protocol}). It reproduces the tool's exact arithmetic per tensor (per row for
\code{qat-convert}). Each iteration moves every mis-exported coordinate, and for 2B-4T's fp32 quantizer every
coordinate inside a relative margin $\mu = 2^{-16}$ of the rounding threshold, by the smallest number of bf16 ULPs, at
most 8, that gives its target code under the current scale, never across zero and at most 16 ULPs from the plain cast.
Where these moves change the tool's scale, single-ULP compensating moves on matching coordinates restore it, bitwise
for the two bf16 exporters and within a relative $\mu/2$ for 2B-4T; a tensor or row whose scale cannot be restored has
that iteration's moves withdrawn, and no accepted iterate may create a new mismatch. The real tool's output is then
checked on every coordinate, for 2B-4T through the compiled reload. At the original remedy studies' endpoints (B2, A1
and C1; Table~\ref{tab:moves}) the projection moved 0.629--1.909\% of the projection coordinates by 1--3 ULPs from
the plain cast. In the recorded runs all 546 projection tensors
converged within two iterations, no margin moves were needed, and the CPU stage took about ten minutes for the three
models together; this records these runs and is not a convergence guarantee. Appendix~\ref{app:protocol} gives the
algorithm as executed (Algorithm~\ref{alg:projection}), what it moved (Table~\ref{tab:moves}) and its portability
checks.

The non-inferiority margin, $\delta = 4$ points, is the median 95\% half-width of the documented-export primaries
(4.199, 5.469 and 4.102 points), rounded down and fixed before launch; it reflects those contrasts' resolution, not an
application tolerance.

\begin{table*}[t]
\centering\scriptsize
\setlength{\tabcolsep}{3pt}
\caption{The two remedies: strict accuracy and cross-entropy against online (95\% paired row bootstrap, 2,000 draws;
$\delta = 4$ points; CE in nats per token).}
\label{tab:remedies}
\begin{adjustbox}{max width=\textwidth}
\begin{tabular}{lrllllll}
\toprule
model & \makecell[l]{online\\strict} & \makecell[l]{fixed $-$\\online} & label & \makecell[l]{projected $-$\\online} & label & \makecell[l]{CE, fixed\\$-$ online} & \makecell[l]{CE, projected\\$-$ online} \\
\midrule
BitCPM-CANN-0.5B (B2) & 36.13\% & \iv{$-0.39$}{$-1.95$, $+1.17$} & non-inferior & \iv{$+0.20$}{$-1.17$, $+1.56$} & non-inferior & \iv{$+0.00006$}{$-0.00006$, $+0.00018$} & \iv{$+0.00013$}{$+0.00001$, $+0.00026$} \\
Falcon-E-1B-Base (A1) & 58.79\% & \iv{$+0.00$}{$-2.73$, $+2.73$} & non-inferior & \iv{$+0.00$}{$-2.54$, $+2.34$} & non-inferior & \iv{$-0.00004$}{$-0.00032$, $+0.00025$} & \iv{$-0.00021$}{$-0.00048$, $+0.00006$} \\
BitNet 2B-4T (C1) & 58.40\% & \iv{$+2.54$}{$+0.59$, $+4.69$} & \makecell[l]{non-inferior,\\detectably above} & \iv{$+0.78$}{$-1.37$, $+2.93$} & non-inferior & \iv{$+0.00017$}{$-0.00015$, $+0.00051$} & \iv{$+0.00018$}{$-0.00013$, $+0.00048$} \\
\bottomrule
\end{tabular}
\end{adjustbox}
\end{table*}

\textbf{What the labels mean.} Each interval's lower bound exceeds $-4$ points, satisfying the predeclared
non-inferiority rule on these rows; this is not a zero-loss guarantee. The intervals still include losses of up to
1.95 and 2.73 points under the fixed export (BitCPM, Falcon-E) and 1.17, 2.54 and 1.37 points under the projection.
Both remedies again met this criterion for the pooled mean over three fixed new trajectories per model, and
separately on the 607 reserve rows (\S\ref{sec:newevidence}).

\textbf{Codes, accuracy and answers.} Both remedies reproduce the online codes on every coordinate, yet the generations
are identical on only part of the 512 rows (Appendix~\ref{app:endpoints}): same codes, same accuracy and same answers
are different claims. Once the codes agree, what remains is the exporter's scale representation,
inference-path arithmetic and the casting of non-projection parameters, which neither experiment isolates. BitNet's
$+2.54$ under the fixed export is detectably positive on these rows, but it is a residual of the complete exported path,
not an established general benefit (Appendix~\ref{app:endpoints}).

\textbf{What each requires.} Both need the online codes as executed, not recomputed on another backend (below). The
fixed export replaces the export
step and, for Falcon-E and BitCPM, writes a convention that differs from the labs' release arithmetic. The projection
keeps the tools unchanged but needs each exporter's exact arithmetic. Where writing the deployed codes directly is permitted, the fixed export is the simpler demonstrated option.
Against the documented state, 2B-4T's strict accuracy changes by $+30.08$ [$+25.00$, $+35.16$] points under the fixed
export, but its last-number accuracy by $-5.08$ [$-9.77$, $-0.39$], so neither remedy supports a
scoring-rule-independent ``accuracy restored'' reading.

\textbf{Cross-backend and thread-reduction sensitivity.} Recomputing the training quantizer on CPU from the same latents
differs from the GPU-captured codes at 59, 427 and 852 coordinates (BitCPM, Falcon-E, 2B-4T), which is why both
remedies take the codes as executed. Scales computed as means can also round differently under another summation order:
for BitCPM, 50 rows whose exact per-row mean lies within $2^{-20}$ of a bf16 rounding midpoint were recorded, since a
host with a different summation order could round them differently, while for Falcon-E, onebitllms' scale was
identical at 1, 4 and 24 CPU threads in all 168 tensors. These are differences between implementations and reduction
orders, not run-to-run nondeterminism within one fixed implementation, and the checks are diagnostics for these tool
versions on this host (Appendix~\ref{app:protocol}). Executed in bitnet.cpp 01eb415 (CPU), the documented export scored 27.54 and 59.38
strict points below the projected export (2B-4T, Falcon-E; last-number $+5.27$ and $-14.45$; code census, intervals and
the cross-implementation comparison in Appendix~\ref{app:e34d}).

\section{bf16 latents during fine-tuning}
\label{sec:training}

The export is not the only place bf16 meets the latents. Fine-tuning can keep the latents themselves in bf16, rounding
optimizer writes to that grid. Under ideal round-to-nearest, a proposed change smaller than half the distance to the
adjacent representable value in its direction is rounded away \citep{zamirai2020bf16}. The on-grid bf16 master-weight
study tests native bf16 and explicit writer emulations prospectively, with its analysis and reporting rules frozen before the run. From the same
released checkpoints as \S\ref{sec:endpoints}'s endpoints (their exact
fp32 upcast for the fp32-storage arms), eleven 300-step GSM8K trainings were evaluated on the same 512 dev rows against that section's fp32 endpoints (2B-4T's C1 and Falcon-E's A1), freshly
re-evaluated; both reproduced their earlier generations row for row.

\textbf{Writers.} Native bf16 (NB) keeps the parameters in bf16 as loaded and trains them with 8-bit AdamW
\citep{dettmers2022optimizers} in bitsandbytes' paged implementation. Three emulations store fp32 values on the bf16 grid and differ only in the
write applied to every trainable parameter after each step: round-to-nearest (RNEe), stochastic rounding (SR;
\citealp{gupta2015limited}) and an error-feedback compensated write (KH, after Kahan summation;
\citealp{zamirai2020bf16}). The emulations isolate the write rule and save no memory. The learning rates are \S\ref{sec:endpoints}'s, $10^{-4}$ for 2B-4T and
$3.125\times10^{-6}$ for Falcon-E; native 2B-4T also ran at $10^{-3}$ and $4\times10^{-3}$.

\textbf{Native writes} (Appendix Table~\ref{tab:training}). In 2B-4T, the native run's evaluation-context codes at the endpoint
equal those at step 0 (zero differences), in this harness. Over its base, strict accuracy rose 55.86
points, last-number accuracy $+4.88$ [$-0.39$, $+10.16$] and cross-entropy $-0.777$ nats per token, with the scorer's
answer-format rate going from 0.027 to 0.965; whatever adaptation occurred went through the amplitudes or the
non-projection parameters, which this design does not separate; a run that trained only the non-projection parameters
matched it in cross-entropy within its declared margin, while last-number accuracy was inconclusive, so the two are
not jointly equivalent, and exclusive learning through those parameters does not follow (Appendix~\ref{app:training}).
Its strict difference from fp32 training is $-0.78$ [$-5.08$, $+3.52$] points, so no strict-accuracy cost is
established, but its cross-entropy is higher by 0.107 [0.100, 0.115] nats per token. In Falcon-E one code changed, an addition in
layer 0; the native run's strict correctness equals its base's on every row, while its last-number accuracy rose
$+2.54$ [$+0.39$, $+4.69$] points. Its exported model scores 0.20\%
strict, $-58.59$ [$-62.89$, $-54.30$] points from fp32 training, and the RNEe emulation scored 0.00\%. This is at a
low, latent-ULP-matched learning rate, and we ran no Falcon-E learning-rate ladder; we do not claim it is the typical
outcome of Falcon-E fine-tuning.

\textbf{Stochastic and compensated writes.} Both emulations changed more codes than their declared movement thresholds
and met the non-inferiority criterion in both models: strict accuracy of the deployed target is non-inferior to the
freshly evaluated fp32 online state at $\delta = 4$ points, family-adjusted over the four claims. For Falcon-E's
exported states, the emulated SR/KH procedure met the declared exported-state criterion. KH on 2B-4T passes narrowly,
with a lower bound of $-3.91$ points. No code count is offered as the cause of an accuracy difference.
Appendix~\ref{app:training} gives further results, including the comparisons with native bf16, and the study's scope.

\section{Implications for deploying ternary fine-tunes}
\label{sec:implications}

\textbf{Check codes, not only loss.} A census of deployed against online codes takes seconds on a CPU, and it
measures an export's code changes directly; we recommend running it before any task evaluation, and monitoring code
changes during fine-tuning, not only the loss: at \S\ref{sec:endpoints}'s learning rates, native bf16 writes left 2B-4T's
endpoint codes equal to its starting codes and changed one Falcon-E code (\S\ref{sec:training}).

\textbf{Export with the quantizer the model was trained under.} Keep the fp32 latents or the codes they produced, and
either write those codes directly (the fixed export) or, if the toolchain must stay unchanged, adjust its inputs (the
projection).

\textbf{Score more than one way.} At the original 2B-4T endpoint, strict accuracy and format adherence fell while
last-number accuracy rose, whereas Falcon-E's and BitCPM's losses appeared under both scoring rules; report both
scoring rules and the format and stop diagnostics, since a single metric would have hidden one or the other. Appendix~\ref{app:endpoints} discusses where else to expect reversion.

\section{Related work}
\label{sec:related}

\textbf{Precision and requantization.} Nearest rounding in bf16 cancels small weight updates, which stochastic
rounding \citep{gupta2015limited} and compensated summation \citep{zamirai2020bf16} counter; we test emulations of
both in ternary fine-tuning (\S\ref{sec:training}), and in the reinforcement-learning configurations that \citet{miahi2026pulse} measured most
per-step updates leave bf16 snapshots unchanged; we measure the analogous one-time cast at export. Requantization can corrupt small
post-training deltas \citep{yu2026daq}, open a deployment gap \citep{wu2026gradcodes}, and delete an adapter merged into
a native 4-bit checkpoint, which Scale-QLoRA avoids by freezing codes and training scales \citep{li2026scaleqlora}.
QATFactory treats the inference engine as the numerical specification and exports without a further lossy conversion
\citep{xu2026qatfactory}; our fixed export is a ternary instance of that principle. Work on latent inertia, threshold
distance, adaptive rounding and training without full-precision masters is discussed in Appendix~\ref{app:selection}.

\textbf{Release artifacts.} The Axolotl/TII guide describes converting trained checkpoints to ternary format without a
before-and-after comparison \citep{axolotlblog}; OpenBMB's report describes the export as saving native ternary tensors
\citep{bitcpmreport}. A public issue and its linked write-up report disagreement between 2B-4T's bf16 master and its packed
checkpoint in sampled tensors \citep{bitnet632,ternarycheck2026}; we give the full census and a fine-tuning
consequence. In a bounded search we found no report of the bf16 export reverting fine-tuning of these models.

\section{Limitations}
\label{sec:limits}

The fine-tuned-endpoint results rest on one recipe family; five trajectories for each headline recipe and one for
every other arm; 512 GSM8K dev prompts from the training split, reused across analyses and for rate selection; and 607
reserve test rows that are not necessarily absent from the base models' pretraining, all under one zero-shot greedy
decoding configuration; greedy answers are sensitive to the batch configuration. The ARC and BoolQ comparisons cover three model/task cases, one trajectory each, on validation rows observed during
training and under one rank-choice measurement convention, and the export-arithmetic training one recipe and rate
(\S\ref{sec:endpoints}). We measure PyTorch bf16 paths and, for the export remedies, one CPU runtime, bitnet.cpp 01eb415 (two models, 512 dev
rows, one trajectory each, timings on a shared host); CANN was not run, and running the exports through onebitllms'
Triton kernels or Microsoft's GPU W2A8 kernels was deferred. The original remedy studies use one existing endpoint per model and one attempt each, which the
new-seed study extends to three fixed new trajectories per model and to the reserve-row evaluation of the original
endpoints; non-inferiority at a
resolution-based margin is not equivalence, and the projection holds only for these tool versions. The on-grid bf16 master-weight study
uses one recipe, one learning rate per model (plus two for 2B-4T) and one trajectory per arm; its SR and KH writers
are emulations without memory savings, and Falcon-E has no learning-rate ladder. A planned benchmark of in-place bf16
stochastic-rounding writes (memory and step time) was dropped under its own pre-declared deadline rule, so we make no
memory or speed claim for SR or KH. The threshold-distance study examines one
released model in depth, with one randomized intervention per model family, first-order scores and an
outcome-dependent support rule. GPU and CPU co-tenancy is disclosed in Appendix~\ref{app:protocol}. Our literature search was bounded, and we make no priority claim.

\section{Conclusion}

The documented bf16 exports of three labs' ternary models can silently revert most of the code changes made by
low-learning-rate fine-tuning, collapsing Falcon-E's and BitCPM's strict-format accuracy and lowering 2B-4T's at the
original low-rate endpoints, at three new seeds per model and on reserve test rows; how much task accuracy is lost
depends on the scoring rule: 2B-4T's last-number accuracy rose detectably at its original endpoints and at two new
seeds, while the third new seed and the reserve-row contrast are unresolved. The effect is recipe-dependent: at Falcon-E's
online-loss-selected rate the export was non-inferior, and after a longer Falcon-E recipe no difference was detected.
Exporting the codes the training quantizer actually produced, or adjusting the inputs so that the unchanged tools emit
them, avoided the strict-accuracy collapse and met the declared, inherited four-point non-inferiority rule on these
rows, for the pooled mean over three fixed new trajectories per model, and separately on the 607 reserve rows. Keeping the latents in bf16 during fine-tuning
raises a related issue: at the tested low learning rates, native bf16 writes left 2B-4T's endpoint codes unchanged and
changed one Falcon-E code, while emulated stochastic and compensated writes changed codes and met the declared criterion.
Fine-tuning pipelines for ternary models should export with the quantizer the model was trained under
and verify the deployed codes against the online ones.

\ifdefined\isarxiv   % named build only: acknowledgments would identify the authors in the double-blind submission
\section*{Acknowledgments}
We thank Bhuvan Gupta (Ofbusiness) for providing the compute infrastructure and for research guidance.
\fi

\begingroup\hexurlbreaks
\bibliography{refs}
\endgroup
\bibliographystyle{mlsys2025}

\section*{AI Usage Disclosure}
\textbf{Human contributions.} The authors developed the export-repair method and designed the experiments. Their contributions included
setting the research agenda and priorities, making scope, launch and protected-data-use decisions, and verifying the work. The authors verified everything reported in this
submission, including the AI-produced code, analyses, references and text.

\textbf{AI assistance.} The authors used Claude Code (Claude Opus 5.5 and Claude Fable 5.1), OpenAI Codex (gpt-6-astra) and Google Gemini 3.1 Pro primarily for iterative coding.
AI generated and iteratively revised implementation and analysis code under the authors' direction.
Additional assistance covered literature searches, pre-registration documentation, manuscript revision and validation.
Codex and Fable performed independent evidence checks; Gemini supported software testing and produced external critiques of the preprint that informed later experiments.

\textbf{Verification and responsibility.} Results come from recorded model runs; synthetic examples were used only for software tests. For pre-registered
experiments, the analysis procedure was set before the runs, with exceptions documented in
Appendix~\ref{app:protocol}. References were checked against primary sources, and automated checks linked
reported numbers to saved records. AI reviews provided additional consistency checks, not human peer review. The
authors directed manuscript revisions and retain full responsibility for the accuracy and integrity of the code,
analyses, references, text and conclusions.

\clearpage
\onecolumn
\setlength{\intextsep}{10pt plus 2pt minus 2pt}   % appendix tables sit in the text ([H])
\appendix

\section{Protocol and provenance}
\label{app:protocol}

\textbf{Declared experiments.} For the prospectively packaged experiments, each was pre-registered before its code was
written and frozen as a package---a manifest of source and record hashes, a spec digest and pinned inputs---after a
build check on training rows, with its analysis code frozen before launch. A controller verified the package at its
phase boundaries, refused to launch if the GPU was occupied at those checks, and recorded each phase's GPU time. The
exceptions below are part of the record.

\textbf{Review.} AI review timing differed across stages. Later packages received separate pre-launch reviews where
recorded; for the two 2B-4T distance interventions the recorded package assessments came after launch, for the second
together with its result review. Result reviews recomputed the primary contrasts from the saved records (AI Usage Disclosure). For the
reserve rows, the result review checked records, provenance and aggregate statistics; it did not rescore answers,
replay the seeded bootstrap or inspect the protected exposure journal. For the other-task and training-forward study, the
result review re-executed the frozen analyzers on CPU, reproducing the analysis and its sentences byte for byte from the
saved scores and generations, and compared the saved training-forward codes and scales with the real tool's saved
output in all 168 projection tensors; it repeated no inference and does not certify invariance to the execution
context. Reporting
corrections were applied in dated successor reports, and frozen records were retained.

\textbf{AI-review provenance.} ``Independent review'' means a pass by a second AI reviewer working from the saved evidence files,
separate from the session that produced the result; these are consistency checks, not human peer
review.\ifarrangement{} The arrangement study's re-execution was
launched by the Claude session under its frozen package and controller on the authors' standing instruction. Its result
review was a substitute independent review by a second AI reviewer (Anthropic Claude, Fable 5.1) while Codex was
unavailable, with Codex's confirmation received on 2026-10-23.\fi{} The bitnet.cpp study's result review was also a
substitute independent review, with Codex's confirmation received on 2026-10-23.

AI assistants searched and summarised the arXiv literature, checked every bibliography entry
against its primary record (arXiv, publisher, repository or model page), and corrected discrepancies.

\textbf{Numbers.} Tables~\ref{tab:app3}--\ref{tab:app10} and \ref{tab:app12}--\ref{tab:app14} are converted cell for cell
from the internal manuscript, whose numbers are checked by script against the accepted runs' evidence files.
Tables~\ref{tab:e34}--\ref{tab:e35states} and \ref{tab:e34d} are generated cell for cell by the scripts
\code{e34\_tables\_to\_tex.py}, \code{e35\_tables\_to\_tex.py} and \code{e34d\_tables\_to\_tex.py} from the hash-pinned
\code{analysis.json} records of the new-seed, rate and reserve-row study, the other-task and training-forward study and
the bitnet.cpp study, and the on-grid study's Tables~\ref{tab:training}--\ref{tab:app33b} by
\code{e33\_tables\_to\_tex.py} from its hash-pinned analysis record and its accepted report's tables.
Tables~\ref{tab:moves} and \ref{tab:intervention} are set in the source, and the checking script matches their rows to
the runs' evidence files; it checks Table~\ref{tab:key}'s registry labels against the same files. In the tables
converted from the manuscript or the on-grid study's records, only the experiment names in captions and headers and
the result labels differ from their sources, and result labels appear in the plain language of Table~\ref{tab:key}
throughout. In the main text, the key results are checked directly against those evidence files, and every other
number is checked for its presence in the internal manuscript. That presence check does not establish the number's
context, which was reviewed editorially.

\textbf{Exceptions, disclosed.}
\begin{itemize}
\item The GSM8K threshold-distance association was found exploratorily; BoolQ and MATH were declared prospectively.
The finite-edit interaction analysis and the Falcon-E export factorial were adaptive.
\item The first 2B-4T distance intervention's input bindings were verified after its verdict, and its probe identity
is unverified; it was accepted as qualified pilot evidence. Its seed-2 repeat was bound before launch.
\item The 2B-4T save-and-reload evaluation is the declared retry of an earlier evaluation, which an out-of-memory
defect had blocked; it evaluated the same retained endpoints, and nothing was retrained.
\item The Falcon-E distance intervention was declared after the Falcon-E observational replication's failure and a
post hoc re-examination of its rule, and its prediction's estimator was chosen from that replication's data. Its
report generator was completed after the launch, and its hash was recorded after qualification, before the primary
analysis existed.
\item Direct code serialization was launched inside a time window that its pre-launch review had excluded.
\item The projection's stage-2 package was built on 2026-10-01 at 15:40:29Z, after the fixed export's and stage 1's
outcomes were known, with the declared design unchanged; its freeze record, which states that time, was written on
2026-10-03 at 06:55:37Z, before the claim.
\item The on-grid bf16 master-weight study's controller died during the run; the interrupted training was retried
under a successor package that raised only the GPU reservation, after an owner decision (Appendix~\ref{app:training}).
\item Free-GPU checks ran at recorded phase boundaries, not continuously. A production job shared the GPU for part of
the evaluation of BitCPM's B2 through OpenBMB's pipeline and for part of the evaluation of BitCPM's fixed export, and
the sampled occupancy log cannot prove exclusivity.
\item During the new-seed, rate and reserve-row study, other CPU work overlapped parts of the run. All ten stage-1
projection records reported four torch CPU threads, and no numerical failure attributable to the overlap appears in
the records; no exclusive-host counterfactual was run. The reserve rows were read once, in a single acquisition.
\item During the other-task and training-forward study, other CPU-only work ran on the host, including another
package's build checks at up to 16 threads. The GPU was free at every recorded phase admission, which records those
admissions rather than continuous surveillance; no numerical failure attributable to the overlap appears in the
records, and no exclusive-host counterfactual was run. Its frozen controller's launch guard matched only an exact
process name that the assistant's executable did not carry, so it could not have fired; the launch was made detached,
and its isolation was verified at launch (parent process ID 1 and a dedicated scope), contemporaneous evidence that
cannot be re-observed after the process exited.
\ifarrangement
\item The arrangement study (internal study S1) was first run outside the packaged pipeline and was then re-executed
under it, after its original results were known; the re-execution is a procedural check, not a blind test. Its rule,
arms, seeds and recipe are those written into S1's plan before S1's first training run; the re-execution's
pre-registration, package (manifest, spec digest, pinned inputs) and analysis code were frozen before its launch, and
a controller ran it with free-GPU checks at phase boundaries. The re-execution's step-zero witness also covers every
non-BitLinear parameter, which S1's did not; its C4 gradient scores are S1's, computed at batch 1 instead of 2. The
re-execution equalled S1's original records in every specified compared field and array. A positive GPU snapshot is
persisted only before the claim; at later phase boundaries the free-GPU check rests on the fail-closed admission gate
and the absence of any pause record.
\fi
\end{itemize}

\textbf{Projection details.} Both remedies need the codes as executed: recomputing them on another backend moved up to
852 threshold decisions (\S\ref{sec:remedies}). Algorithm~\ref{alg:projection} follows the executed implementation. Its portability checks
are diagnostics, not guarantees across arbitrary kernels or versions. For Falcon-E, onebitllms' scale was identical at
1, 4 and 24 CPU threads in all 168 tensors, and the implementation includes a safeguard that keeps a tensor's exact mean
magnitude at least a relative $2^{-20}$ away from a bf16 rounding midpoint. For BitCPM, 50 rows whose exact per-row mean
lies within $2^{-20}$ of a bf16 rounding midpoint were recorded, since a host with a different summation order could
round them differently. For 2B-4T, every fp32 product stays at least the relative margin $\mu$ clear of the threshold,
and the compiled GPU census passed on every coordinate in the build check and in the run.

\begin{algorithm}[H]
\caption{The ULP-compensated input projection as executed, per tensor; per row for \code{qat-convert}.}
\label{alg:projection}
\begin{algorithmic}
\STATE {\bfseries Input:} online codes $T$ as executed; the plain cast $v_0 = \mathrm{bf16}(W)$; the exporter's exact
scale $S(v)$ and codes $C(v, s)$; margin $\mu$ ($2^{-16}$ for 2B-4T, else 0)
\STATE $v \leftarrow v_0$
\FOR{$\mathit{it} = 1$ {\bfseries to} $16$}
\STATE $s \leftarrow S(v)$; \ $M \leftarrow \{i$ not already a residual$: C(v,s)_i \neq T_i$, or $i$ inside the
margin$\}$
\IF{$M = \emptyset$}
\STATE {\bfseries stop}
\ENDIF
\STATE for each $i \in M$: the smallest $k \le 8$ such that moving $|v_i|$ by $k$ ULPs toward $T_i$'s side (not across
zero, within 16 ULPs of $v_0$) gives $C(\cdot, s)_i = T_i$ outside the margin; if none, $i$ is a residual
\IF{the moves change the scale (its bf16 bits; beyond a relative $\mu/2$ for 2B-4T)}
\STATE add single-ULP moves on matched, unmoved coordinates, by descending $|v|$: the shortest prefix, at most 1\% of
the group, that restores the scale; if none does, withdraw the group's moves and stop the group
\ENDIF
\STATE accept the iterate only if it creates no new mismatch; otherwise withdraw that group's moves and stop the group
\ENDFOR
\STATE record residual mismatches and margin failures, with reasons
\STATE run the real tool on $v$ (2B-4T: the compiled reload) and require its codes to equal $T$ everywhere
\end{algorithmic}
\end{algorithm}
\begin{table}[H]
\centering\small
\setlength{\tabcolsep}{2.5pt}
\caption{What the projection moved: code-correcting moves, scale-compensating moves on coordinates whose codes already
matched, and the bf16 ULP displacement from the plain cast. Every tensor converged within two iterations.}
\label{tab:moves}
\begin{adjustbox}{max width=\textwidth}
\begin{tabular}{lrrrrl}
\toprule
model & coordinates & correcting & compensating & moved & ULP: count \\
\midrule
BitCPM-CANN-0.5B & 358,612,992 & 5,985,171 & 277 & 1.669\% & 1: 5,667,377; 2: 318,066; 3: 5 \\
Falcon-E-1B-Base & 1,585,446,912 & 30,266,748 & 0 & 1.909\% & 1: 27,418,368; 2: 2,848,380 \\
BitNet 2B-4T & 2,084,044,800 & 12,651,712 & 449,625 & 0.629\% & 1: 13,101,337 \\
\bottomrule
\end{tabular}
\end{adjustbox}
\end{table}

\textbf{What the projection moves.} It moved 1.669\%, 1.909\% and 0.629\% of the projection coordinates (BitCPM,
Falcon-E, 2B-4T) by 1--3 ULPs from the plain bf16 cast, including scale-compensating moves on coordinates whose codes
already matched (277 in BitCPM, 449,625 in 2B-4T). Every code the documented exports got wrong at these endpoints was
corrected by such a move. The median relative L2 distance to the fp32 latents rises from 2.97, 3.01 and
2.53$\times10^{-4}$ for the plain cast to 4.34, 4.25 and 4.03$\times10^{-4}$. Global minimality is not established.

\textbf{Compute.} All GPU runs used one NVIDIA RTX 4090; the bitnet.cpp study ran on CPU. A 300-step 2B-4T fine-tune
takes about 1,800 GPU-seconds, and a Falcon-E fine-tune about 1,100.

\textbf{Reproducibility key.} Table~\ref{tab:key} maps each study's name in this paper to the project's internal
registry, gives its result labels in this paper beside the registry's own labels, and names the evidence files that
its numbers are checked against.

\begin{table}[tbp]
\centering\scriptsize
\setlength{\tabcolsep}{3pt}
\renewcommand{\arraystretch}{1.12}
\caption{The reproducibility key. Paths are relative to the project root; the files in \nolinkurl{reports/paper/} are
the paper's hash-pinned evidence digests, with the key read in parentheses, and they pin the runs' analysis records by
sha256.}
\label{tab:key}
\begin{tabular}{>{\raggedright\arraybackslash}p{1.55in}>{\raggedright\arraybackslash}p{0.88in}>{\raggedright\arraybackslash}p{2.25in}>{\raggedright\arraybackslash}p{1.66in}}
\toprule
name in this paper & registry & label in this paper: registry label & evidence \\
\midrule
release census, tie analysis and exporter reproduction (\S\ref{sec:release}) & release audit (no number) & -- & \nolinkurl{reports/paper/release_audit_evidence.json} \\
release-loss comparison on C4 (Table~\ref{tab:app3}) & Experiment 14 & deployed worse: \code{DEPLOYED\_WORSE}; deployed better: \code{DEPLOYED\_BETTER}; no detectable difference: \code{NO\_DETECTABLE\_DIFFERENCE} & the same (\code{e14}) \\
BitCPM precision contrast (Table~\ref{tab:app4}) & Experiment 16 & deployed better: \code{DEPLOYED\_BETTER} & the same (\code{e16}) \\
Falcon-E learning-rate arms A1--A4 (\S\ref{sec:endpoints}; Tables~\ref{tab:app5}, \ref{tab:app6}) & Experiment 17 & export costs: \code{EXPORT\_COSTS}; no detectable difference: \code{NO\_DETECTABLE\_DIFFERENCE}; export helps: \code{EXPORT\_HELPS} & \nolinkurl{reports/paper/export_ft_evidence.json} (\code{e17}) \\
Falcon-E export factorial (\S\ref{sec:endpoints}) & Experiment 18 & -- & the same (\code{e18}) \\
Falcon-E rate ladder B1--B4 (Table~\ref{tab:app7}) & Experiment 19 & export costs: \code{EXPORT\_COSTS} & the same (\code{e19}) \\
Falcon-E learning-rate arms scored for accuracy (Table~\ref{tab:app8}) & Experiment 20 & export lowers accuracy: \code{EXPORT\_LOWERS\_ACCURACY}; no detectable difference: \code{NO\_DETECTABLE\_DIFFERENCE} & the same (\code{e20}) \\
2B-4T save-and-reload evaluation C1--C2 (Table~\ref{tab:app9}) & Experiment 22, the declared retry of Experiment 21 & save lowers accuracy: \code{SAVE\_LOWERS\_ACCURACY} & \nolinkurl{reports/paper/crosslab_evidence.json} (\code{e22}) \\
BitCPM through OpenBMB's pipeline, low rate, B1--B2 (Table~\ref{tab:app10}) & Experiment 23 & export lowers accuracy: \code{EXPORT\_LOWERS\_ACCURACY} & the same (\code{e23}) \\
BitCPM through OpenBMB's pipeline at 32 times the rate, B3 (Table~\ref{tab:endpoints}) & Experiment 24 & no detectable difference: \code{NO\_DETECTABLE\_DIFFERENCE} & the same (\code{e24}) \\
direct code serialization, the fixed export (\S\ref{sec:remedies}) & Experiment 26 & non-inferior: \code{FIX\_NONINFERIOR}; detectably above: \code{DETECTABLY\_ABOVE} & \nolinkurl{reports/paper/remedies_evidence.json} (\code{e26}) \\
ULP-compensated input projection, the projection (\S\ref{sec:remedies}) & Experiment 27 & non-inferior: \code{PROJ\_NONINFERIOR} & the same (\code{e27}); \nolinkurl{data/slm_runs/_experiment27/stage1/} \code{\{O,F,M\}/attempt\_1/record.json} \\
on-grid bf16 master-weight study (\S\ref{sec:training}, Appendix~\ref{app:training}) & Experiment 33 & endpoint codes frozen, not frozen: \code{FROZEN\_ENDPOINT}, \code{NOT\_FROZEN}; non-inferior: \code{NONINFERIOR}; method-success rule met: \code{METHOD\_SUCCESS}; training-context codes frozen, not frozen: \code{GRAD\_PREFIX\_FROZEN}, \code{GRAD\_PREFIX\_NOT\_FROZEN}; the onset forecast holds: \code{ONSET\_FORECAST\_HOLDS}; the cast left codes unchanged: \code{CAST\_CODE\_SAFE}; within $\pm0.01$ nats, inconclusive at $\pm2$ points, not jointly equivalent: \code{EQUIVALENT}, \code{INCONCLUSIVE}, \code{NOT\_JOINT\_EQUIVALENT}; reproduced row for row: \code{REPLAY\_EXACT} & \nolinkurl{data/slm_runs/_experiment33/analysis/run_001_analysis.json}; \nolinkurl{.../result_tables_rev3.json} \\
public fine-tune census (Appendix~\ref{app:training}) & audit L, completeness binding & -- & \nolinkurl{research_round2/scouting/audit_L/completeness/binding.json} \\
\bottomrule
\end{tabular}
\end{table}
\begin{table}[tbp]
\centering\scriptsize
\setlength{\tabcolsep}{3pt}
\renewcommand{\arraystretch}{1.12}
\addtocounter{table}{-1}
\caption{The reproducibility key, continued.}
\label{tab:keycont}
\begin{tabular}{>{\raggedright\arraybackslash}p{1.55in}>{\raggedright\arraybackslash}p{0.88in}>{\raggedright\arraybackslash}p{2.25in}>{\raggedright\arraybackslash}p{1.66in}}
\toprule
name in this paper & registry & label in this paper: registry label & evidence \\
\midrule
threshold-distance association on GSM8K, two seeds (\S\ref{sec:selection}, Table~\ref{tab:app12}) & Experiment 6 (exploratory) & survives: \code{SURVIVES} & \nolinkurl{reports/paper/claimB_evidence.json} (\code{endpoints}) \\
its BoolQ and MATH endpoints (declared) & Experiments 7 and 8 & reported as contrasts; registry: \code{GRID\_REPLICATED} & the same (\code{endpoints}) \\
first 2B-4T distance intervention (Tables~\ref{tab:intervention}, \ref{tab:app13}) & Experiment 9 & supported: \code{SUPPORTED} & \nolinkurl{data/slm_runs/_experiment9/runs/run_001/analysis.json} \\
its seed-2 repeat & Experiment 9R & supported: \code{SUPPORTED} & \nolinkurl{.../_experiment9/runs/seed2_001/analysis.json} \\
finite-edit reversion test (Appendix~\ref{app:selection}) & Experiment 10 & reported as a contrast; registry: \code{FAR\_CARRIES\_MORE} & \nolinkurl{reports/paper/claimB_evidence.json} (\code{functional\_and\_transfer}) \\
finite-edit interaction analysis (Table~\ref{tab:app14}) & Experiment 12 & reported as contrasts; registry: \code{FAR\_INTERACTION\_GREATER} & the same \\
Falcon-E observational replication (Appendix~\ref{app:selection}) & Experiment 13 & not replicated: \code{NOT\_REPLICATED} & \nolinkurl{data/slm_runs/_experiment13/runs/run_001/analysis.json} \\
Falcon-E distance intervention (Table~\ref{tab:intervention}) & Experiment 25 & supported: \code{SUPPORTED} & \nolinkurl{reports/paper/remedies_evidence.json} (\code{e25}) \\
\ifarrangement
the arrangement study (Appendix~\ref{app:arrangement}) & internal study S1, re-executed under the pipeline & its own frozen labels, used verbatim: equivalent, effect, inconclusive (Stage B); information present, equivalent to the null, inconclusive (Stage A) & \nolinkurl{data/slm_runs/_arrangement/runs/run_001/analysis.json}; \nolinkurl{.../sentences.md} \\
\fi
new-seed, rate and reserve-row study: Falcon-E rates (Part A; Table~\ref{tab:e34}) & Experiment 34, Part A & non-inferior at the tuned rate: \code{EXPORT\_NONINFERIOR\_AT\_TUNED\_RATE}; export lowers accuracy: \code{EXPORT\_LOWERS\_ACCURACY}; lower, more than 4 points below: \code{LOWER + DEPLOY\_TUNED\_INFERIOR} & \nolinkurl{data/slm_runs/_experiment34/runs/run_001/analysis.json}; \nolinkurl{.../sentences.md} \\
its new seeds, 2B-4T rates (A-BN) and longer recipe (A-long) & Experiment 34, Part B & lower at all three seeds: \code{SEEDS\_ALL\_LOWER}; pooled non-inferior: \code{FIX\_POOLED\_NONINFERIOR}, \code{PROJ\_POOLED\_NONINFERIOR}; lower: \code{LOWER}; no detectable difference: \code{NO\_DETECTABLE\_DIFFERENCE} & the same \\
its reserve rows (Part C) & Experiment 34, Part C & same direction as dev: \code{HELD\_OUT\_SAME\_DIRECTION}; non-inferior: \code{FIX\_NONINFERIOR}, \code{PROJ\_NONINFERIOR}; last-number also lower, unresolved: \code{STRICT\_AND\_LAST\_NUMBER\_LOWER}, \code{STRICT\_LOWER\_LAST\_NUMBER\_UNRESOLVED}; boundary flag: \code{BOUNDARY} & the same \\
other-task and training-forward study: other tasks (Part T; Table~\ref{tab:e35}) & Experiment 35, Part T & export lowers accuracy: \code{EXPORT\_LOWERS\_ACCURACY}; no detectable difference: \code{NO\_DETECTABLE\_DIFFERENCE}; more than 4 points below, inconclusive at $\delta = 4$ pp: \code{EXPORT\_INFERIOR}, \code{EXPORT\_INCONCLUSIVE}; non-inferior: \code{FIX\_NONINFERIOR}, \code{PROJ\_NONINFERIOR}; fine-tuning gain established (qualifier): \code{GAIN\_ESTABLISHED}; boundary flag: \code{BOUNDARY} & \nolinkurl{data/slm_runs/_experiment35/runs/run_001/analysis.json}; \nolinkurl{.../sentences.md} \\
its export-arithmetic training forward (Part DA) & Experiment 35, Part DA & non-inferior: \code{DA\_NONINFERIOR}; no detectable difference: \code{NO\_DETECTABLE\_DIFFERENCE}; higher: \code{HIGHER} & the same \\
ternary exports in bitnet.cpp 01eb415 on CPU (R1; Appendix~\ref{app:e34d}) & Experiment 34-D & documented below projected: \code{DOCUMENTED\_BELOW\_REMEDY}; carries the projected codes: \code{RUNTIME\_CARRIES\_PROJECTED\_CODES}; no detectable difference and non-inferior to the PyTorch evaluation: \code{NO\_DETECTABLE\_DIFFERENCE}, \code{RUNTIME\_REMEDY\_NONINFERIOR}; secondary: \code{LOWER}, \code{NO\_DETECTABLE\_DIFFERENCE} & \nolinkurl{data/slm_runs/_experiment34d/runs/run_001/phases/D.analyze/attempt_2/analysis.json}; \nolinkurl{.../sentences.md} \\
\bottomrule
\end{tabular}
\end{table}

\section{Additional release-audit results}
\label{app:release}
\textbf{The tie rule} (\S\ref{sec:release}). Rounding ties away from zero instead turns the error mostly into additions,
so changing the tie rule changes the dominant direction without removing the precision-induced disagreement.

\textbf{2B-4T's hidden codes and fine-tuning} (\S\ref{sec:release}). 2B-4T's disagreements matter for fine-tuning: about
41\% of all code changes made by fine-tuning 2B-4T from its bf16 master land on these coordinates and afterwards agree
with the packed code. Because the pre-release latents are not observed, this agreement is operational, not a claim of
restoration.

\textbf{The loss consequence depends on the release.} In a declared comparison on C4 \citep{raffel2020c4} (validation
shard 00001, 1,024 documents, one common forward with int8 activations), the deployed codes and scales raise
cross-entropy over fp32 quantization of the released latents by $+0.0125$ [$+0.0114$, $+0.0137$] nats per token for
Falcon-E-1B-Base and $+0.0077$ [$+0.0066$, $+0.0087$] for 3B-Base, lower it for 2B-4T ($-0.0053$ [$-0.0067$,
$-0.0039$]), and show no detectable difference for the two Instruct releases (Table~\ref{tab:app3}). In a
separate precision contrast on the same BitCPM-CANN latents, the release's bf16 output has lower loss than an fp32
counterfactual ($-0.035544$ [$-0.037267$, $-0.033914$] for 0.5B and $-0.006814$ [$-0.007933$, $-0.005625$] for 3B), and
rounding ties away from zero is worse than both (Table~\ref{tab:app4}). The two comparisons measure different estimands
in different forwards, so we infer no rule across releases: the practical point is consistency with the quantizer a
model was trained under, not that fp32 is better.

Table~\ref{tab:app3} gives the C4 comparison of \S\ref{sec:release} (the release-loss comparison): A is fp32 quantization of the
released latents, B the deployed codes and scales, C the deployed codes at A's scale and D A's codes at the deployed
scale, so that $B-A = (C-A) + (D-A) + I$, with the interaction $I$ reported descriptively. The Falcon-E intervals share
documents and bootstrap draws, so they are correlated, and no detectable difference is not equivalence.
Table~\ref{tab:app4} is the BitCPM precision contrast: B is the release's own bf16 arithmetic on the
released latents, which reproduces the deployed weights, A the fp32-arithmetic counterfactual on the same latents, C and
D swap codes and magnitudes as above, and E rounds ties away from zero at the bf16 product and magnitude.
\begin{table}[H]
\centering\scriptsize
\setlength{\tabcolsep}{3pt}
\caption{B $-$ A in nats per predicted token [95\% document bootstrap]. C $-$ A swaps the codes at A's scale, and D $-$ A swaps the scale at A's codes.}
\label{tab:app3}
\begin{adjustbox}{max width=\textwidth}
\begin{tabular}{llllll}
\toprule
release & label & B $-$ A & C $-$ A & D $-$ A & I \\
\midrule
Falcon-E-1B-Base & deployed worse & \makecell[l]{+0.0125\\{[+0.0114, +0.0137]}} & \makecell[l]{+0.0125\\{[+0.0113, +0.0136]}} & \makecell[l]{$-$0.0002\\{[$-$0.0005, +0.0000]}} & +3.0e-04 \\
Falcon-E-1B-Instruct & no detectable difference & \makecell[l]{$-$0.0006\\{[$-$0.0018, +0.0007]}} & \makecell[l]{$-$0.0004\\{[$-$0.0017, +0.0008]}} & \makecell[l]{+0.0000\\{[$-$0.0002, +0.0003]}} & $-$1.8e-04 \\
BitNet b1.58 2B-4T & deployed better & \makecell[l]{$-$0.0053\\{[$-$0.0067, $-$0.0039]}} & \makecell[l]{$-$0.0053\\{[$-$0.0067, $-$0.0039]}} & \makecell[l]{+0.0000\\{[+0.0000, +0.0000]}} & +0.0e+00 \\
Falcon-E-3B-Base & deployed worse & \makecell[l]{+0.0077\\{[+0.0066, +0.0087]}} & \makecell[l]{+0.0079\\{[+0.0069, +0.0090]}} & \makecell[l]{$-$0.0001\\{[$-$0.0004, +0.0001]}} & $-$1.2e-04 \\
Falcon-E-3B-Instruct & no detectable difference & \makecell[l]{+0.0009\\{[$-$0.0000, +0.0019]}} & \makecell[l]{+0.0011\\{[+0.0002, +0.0021]}} & \makecell[l]{$-$0.0002\\{[$-$0.0004, +0.0001]}} & $-$1.0e-06 \\
\bottomrule
\end{tabular}
\end{adjustbox}
\end{table}

\begin{table}[H]
\centering\scriptsize
\setlength{\tabcolsep}{3pt}
\caption{The BitCPM precision contrast, nats per predicted token [95\% document bootstrap].}
\label{tab:app4}
\begin{adjustbox}{max width=\textwidth}
\begin{tabular}{llllllll}
\toprule
release & label & B $-$ A & C $-$ A & D $-$ A & E $-$ A & E $-$ B & I \\
\midrule
BitCPM-CANN-0.5B & deployed better & \makecell[l]{$-$0.035544\\{[$-$0.037267, $-$0.033914]}} & \makecell[l]{$-$0.035645\\{[$-$0.037337, $-$0.034011]}} & \makecell[l]{+0.000131\\{[+0.000035, +0.000226]}} & \makecell[l]{+0.024380\\{[+0.022860, +0.025848]}} & \makecell[l]{+0.059923\\{[+0.057852, +0.062146]}} & $-$3.01e-05 \\
BitCPM-CANN-3B & deployed better & \makecell[l]{$-$0.006814\\{[$-$0.007933, $-$0.005625]}} & \makecell[l]{$-$0.006617\\{[$-$0.007754, $-$0.005409]}} & \makecell[l]{$-$0.000402\\{[$-$0.000508, $-$0.000290]}} & \makecell[l]{+0.003834\\{[+0.002750, +0.004922]}} & \makecell[l]{+0.010648\\{[+0.009151, +0.012184]}} & +2.05e-04 \\
\bottomrule
\end{tabular}
\end{adjustbox}
\end{table}

\section{Additional fine-tuned-endpoint results}
\label{app:endpoints}
Tables~\ref{tab:app5}--\ref{tab:app10} give the per-study detail behind \S\ref{sec:endpoints}. Arm labels are local to
each study (Table~\ref{tab:key}): A1--A4 are the Falcon-E learning-rate arms (Table~\ref{tab:app8} scores the same
states for accuracy), B1--B4 in Table~\ref{tab:app7} are the Falcon-E rate-ladder arms, C1--C2 are the 2B-4T
save-and-reload trajectories, and B1--B2 in Table~\ref{tab:app10} are BitCPM's low-rate arms through OpenBMB's pipeline. The three steps of the declared path online $\to$
\code{cast\_Ptrained} $\to$ \code{export\_Ptrained} $\to$ export are the projection cast, the export arithmetic given
the cast latents, and the non-projection cast.

\textbf{Where to expect reversion.} Reversion is expected wherever fine-tuning leaves changed latents inside their
starting values' bf16 rounding cells. Low learning rates make that likely, but it is a property of the endpoint:
long runs can also end near a boundary. In the reinforcement-learning configurations that \citet{miahi2026pulse}
measured, about 99\% of per-step updates leave the bf16 weights unchanged between successive snapshots; that
concerns changes between snapshots, not the permanent loss of accumulated fp32 updates. We did not test RL
fine-tuning of ternary models, and its exposure to export reversion is a hypothesis.

\textbf{The factorial.} A $2\times2\times2$ factorial at the low-rate endpoint, with the non-projection
parameters fixed and the same endpoint and dev rows, crosses the export's two code-change groups, reverted additions
(F1) and newly zeroed codes (F2), with a substitution of the amplitudes. F1 and F2 interact strongly:
$+0.27123$ [$+0.26188$, $+0.28077$] nats per token at the online amplitudes and $+0.27081$ [$+0.26133$, $+0.28051$] at
the export amplitudes, while all four conditional amplitude contrasts include zero.
F1 is not the cast and F2 is not the arithmetic; neither these contrasts nor the ordered-path increments are shares
of the cost.

\textbf{The Falcon-E rate sequence} (\S\ref{sec:endpoints}; Table~\ref{tab:app7}). The cost recurs on a second
trajectory at the same rate ($+0.41401$ [$+0.40139$, $+0.42714$]); no replication rule was declared, and two
trajectories do not measure seed variance. Over the same-seed rates, the
export's cross-entropy cost is $+0.09436$ at $6.25\times10^{-6}$, $+0.01524$ at $10^{-5}$ and $+0.00264$ at
$2\times10^{-5}$ nats per token.

\textbf{BitCPM at a higher rate} (\S\ref{sec:endpoints}; Table~\ref{tab:endpoints}, B3). At 32 times the rate the same
pipeline leaves no detectable strict-accuracy change and a small cross-entropy cost; the cast then returns only 4.0\% of
14,972,249 changes.

\textbf{The remedies' codes and answers} (\S\ref{sec:remedies}). Both remedies reproduce the online codes on every
coordinate---358,612,992, 1,585,446,912 and 2,084,044,800 codes---read from the saved files or the tools' own output
and, for 2B-4T's projection, from the compiled quantizer on the GPU. The generations are nevertheless identical on only
395, 266 and 314 of 512 rows under the fixed export, and 390, 281 and 305 under the projection. The projected 2B-4T
model lies $-1.76$ [$-4.10$, $+0.59$] points from the fixed export, and the two are not ranked. BitCPM's projected
cross-entropy is the one detectable CE difference, 0.03\% higher than online.

\begin{table}[H]
\centering\scriptsize
\setlength{\tabcolsep}{3pt}
\caption{The Falcon-E learning-rate arms, nats per target token [95\% row bootstrap]. The declared path online $\to$ cast\_Ptrained $\to$ export\_Ptrained $\to$ export splits each primary exactly, in that order.}
\label{tab:app5}
\begin{adjustbox}{max width=\textwidth}
\begin{tabular}{llllllll}
\toprule
arm & master & lr & label & export $-$ online & projection cast & export arithmetic, given cast latents & non-projection cast \\
\midrule
A1 & fp32 & 3.125e-06 & export costs & \makecell[l]{+0.41513\\{[+0.40251, +0.42914]}} & \makecell[l]{+0.38053\\{[+0.36875, +0.39357]}} & \makecell[l]{+0.03412\\{[+0.03191, +0.03645]}} & \makecell[l]{+0.00048\\{[+0.00001, +0.00094]}} \\
A2 & bf16 & 0.0005 & export costs & \makecell[l]{+0.00111\\{[+0.00037, +0.00186]}} & \makecell[l]{+0.00000\\{[+0.00000, +0.00000]}} & \makecell[l]{+0.00111\\{[+0.00037, +0.00186]}} & \makecell[l]{+0.00000\\{[+0.00000, +0.00000]}} \\
A3 & fp32 & 0.0005 & no detectable difference & \makecell[l]{+0.00011\\{[$-$0.00041, +0.00065]}} & \makecell[l]{$-$0.00032\\{[$-$0.00069, +0.00003]}} & \makecell[l]{+0.00047\\{[$-$0.00005, +0.00102]}} & \makecell[l]{$-$0.00004\\{[$-$0.00023, +0.00014]}} \\
A4 & fp32 & 5e-05 & export helps & \makecell[l]{$-$0.00147\\{[$-$0.00256, $-$0.00034]}} & \makecell[l]{+0.00017\\{[$-$0.00042, +0.00079]}} & \makecell[l]{$-$0.00149\\{[$-$0.00258, $-$0.00040]}} & \makecell[l]{$-$0.00015\\{[$-$0.00036, +0.00005]}} \\
\bottomrule
\end{tabular}
\end{adjustbox}
\end{table}

\begin{table}[H]
\centering\scriptsize
\setlength{\tabcolsep}{3pt}
\caption{Where each endpoint's code additions sit (descriptive; post hoc summaries of the Falcon-E learning-rate arms' declared margin histograms and flip decompositions). The near share is the fraction of additions within 0.25 bf16 ULP above the fp32 threshold, boundary cases included. The reverted shares are the fractions the cast and the export return to the base code. The continuous margins cannot be recomputed independently without the unsaved fp32 masters; codex checked their provenance and internal consistency.}
\label{tab:app6}
\begin{adjustbox}{max width=\textwidth}
\begin{tabular}{lllll}
\toprule
arm & additions & near share & reverted by the cast & reverted by the export \\
\midrule
A1 & 4,055,607 & 91.2\% & 96.3\% & 98.3\% \\
A2 & 94,497,902 & 0.9\% & 0.0\% & 3.6\% \\
A3 & 100,821,391 & 1.2\% & 0.6\% & 3.0\% \\
A4 & 33,248,011 & 13.1\% & 7.6\% & 28.1\% \\
\bottomrule
\end{tabular}
\end{adjustbox}
\end{table}

\begin{table}[H]
\centering\scriptsize
\setlength{\tabcolsep}{3pt}
\caption{The Falcon-E rate ladder, nats per target token [95\% row bootstrap], along Table~\ref{tab:app5}'s declared path.}
\label{tab:app7}
\begin{adjustbox}{max width=\textwidth}
\begin{tabular}{llllllll}
\toprule
arm & lr & seed & export $-$ online & label & cast\_Ptrained $-$ online & export\_Ptrained $-$ cast\_Ptrained & export $-$ export\_Ptrained \\
\midrule
B1 & 3.125e-06 & 2026092301 & \makecell[l]{+0.41401\\{[+0.40139, +0.42714]}} & export costs & \makecell[l]{+0.38345\\{[+0.37159, +0.39566]}} & \makecell[l]{+0.02962\\{[+0.02746, +0.03192]}} & \makecell[l]{+0.00094\\{[+0.00045, +0.00142]}} \\
B2 & 6.25e-06 & 2026091700 & \makecell[l]{+0.09436\\{[+0.09022, +0.09852]}} & export costs & \makecell[l]{+0.04578\\{[+0.04291, +0.04864]}} & \makecell[l]{+0.04881\\{[+0.04666, +0.05081]}} & \makecell[l]{$-$0.00022\\{[$-$0.00057, +0.00010]}} \\
B3 & 1e-05 & 2026091700 & \makecell[l]{+0.01524\\{[+0.01306, +0.01736]}} & export costs & \makecell[l]{$-$0.00120\\{[$-$0.00212, $-$0.00031]}} & \makecell[l]{+0.01633\\{[+0.01447, +0.01811]}} & \makecell[l]{+0.00010\\{[$-$0.00014, +0.00036]}} \\
B4 & 2e-05 & 2026091700 & \makecell[l]{+0.00264\\{[+0.00118, +0.00405]}} & export costs & \makecell[l]{+0.00064\\{[$-$0.00016, +0.00141]}} & \makecell[l]{+0.00223\\{[+0.00077, +0.00370]}} & \makecell[l]{$-$0.00023\\{[$-$0.00044, $-$0.00001]}} \\
\bottomrule
\end{tabular}
\end{adjustbox}
\end{table}

\begin{table}[H]
\centering\scriptsize
\setlength{\tabcolsep}{3pt}
\caption{The Falcon-E learning-rate arms scored for accuracy: greedy accuracy on the 512 dev prompts (percentage points; 95\% paired row bootstrap).}
\label{tab:app8}
\begin{adjustbox}{max width=\textwidth}
\begin{tabular}{lllllllll}
\toprule
arm & master & lr & online strict / last-number & export strict / last-number & export $-$ online, strict & label & export $-$ online, last-number & format online $\to$ export \\
\midrule
A1 & fp32 & 3.125e-06 & 58.79\% / 58.79\% & 0.78\% / 47.07\% & \makecell[l]{$-$58.01\\{[$-$62.30, $-$53.91]}} & export lowers accuracy & \makecell[l]{$-$11.72\\{[$-$16.99, $-$6.64]}} & 97.1\% $\to$ 21.7\% \\
A2 & bf16 & 0.0005 & 13.48\% / 13.67\% & 11.72\% / 11.91\% & \makecell[l]{$-$1.76\\{[$-$3.71, +0.00]}} & no detectable difference & \makecell[l]{$-$1.76\\{[$-$3.71, +0.00]}} & 87.3\% $\to$ 88.5\% \\
A3 & fp32 & 0.0005 & 15.82\% / 16.21\% & 16.21\% / 16.41\% & \makecell[l]{+0.39\\{[$-$1.56, +2.34]}} & no detectable difference & \makecell[l]{+0.20\\{[$-$1.76, +2.15]}} & 91.0\% $\to$ 92.4\% \\
A4 & fp32 & 5e-05 & 40.04\% / 40.04\% & 37.70\% / 37.50\% & \makecell[l]{$-$2.34\\{[$-$5.47, +0.78]}} & no detectable difference & \makecell[l]{$-$2.54\\{[$-$5.47, +0.39]}} & 94.9\% $\to$ 93.9\% \\
\bottomrule
\end{tabular}
\end{adjustbox}
\end{table}

\begin{table}[H]
\centering\scriptsize
\setlength{\tabcolsep}{3pt}
\caption{The 2B-4T save-and-reload evaluation: greedy accuracy and dev CE, online and reloaded (95\% paired row bootstrap).}
\label{tab:app9}
\begin{adjustbox}{max width=\textwidth}
\begin{tabular}{llllllll}
\toprule
arm & online strict / last-number & reload\_user strict / last-number & reload\_user $-$ online, strict & label & reload\_user $-$ online, last-number & CE, nats/token & format online $\to$ reload\_user \\
\midrule
C1 & 58.40\% / 58.20\% & 30.86\% / 65.82\% & \makecell[l]{$-$27.54\\{[$-$33.01, $-$22.07]}} & save lowers accuracy & \makecell[l]{+7.62\\{[+2.54, +12.11]}} & \makecell[l]{+0.30198\\{[+0.28990, +0.31520]}} & 98.4\% $\to$ 41.4\% \\
C2 & 58.98\% / 58.98\% & 35.35\% / 66.60\% & \makecell[l]{$-$23.63\\{[$-$28.71, $-$18.55]}} & save lowers accuracy & \makecell[l]{+7.62\\{[+2.54, +12.70]}} & \makecell[l]{+0.29636\\{[+0.28450, +0.30898]}} & 98.2\% $\to$ 45.1\% \\
\bottomrule
\end{tabular}
\end{adjustbox}
\end{table}

\begin{table}[H]
\centering\scriptsize
\setlength{\tabcolsep}{3pt}
\caption{BitCPM through OpenBMB's pipeline: greedy accuracy and dev CE, online and deployed (95\% paired row bootstrap). The last two columns are the census: the share of fine-tuning's code changes the bf16 cast returns to the start code, and the codes in which the served model differs from the published checkpoint.}
\label{tab:app10}
\begin{adjustbox}{max width=\textwidth}
\begin{tabular}{lllllllll}
\toprule
arm & online strict / last-number & deployed strict / last-number & deployed $-$ online, strict & label & deployed $-$ online, last-number & CE, nats/token & cast returns & served codes differing \\
\midrule
B1 & 33.98\% / 33.98\% & 0.98\% / 12.11\% & \makecell[l]{$-$33.01\\{[$-$37.11, $-$28.71]}} & export lowers accuracy & \makecell[l]{$-$21.88\\{[$-$26.37, $-$17.58]}} & \makecell[l]{+0.21367\\{[+0.20582, +0.22180]}} & 95.3\% & 98,887 (8.8\%) \\
B2 & 36.13\% / 36.13\% & 0.39\% / 13.09\% & \makecell[l]{$-$35.74\\{[$-$40.04, $-$31.84]}} & export lowers accuracy & \makecell[l]{$-$23.05\\{[$-$27.54, $-$18.95]}} & \makecell[l]{+0.21283\\{[+0.20486, +0.22082]}} & 95.3\% & 99,080 (8.8\%) \\
\bottomrule
\end{tabular}
\end{adjustbox}
\end{table}

\textbf{The new-seed, rate and reserve-row study} (\S\ref{sec:newevidence}). Tables~\ref{tab:e34} and \ref{tab:e34rem} list its 23 primary
contrasts (the second with each pooled remedy row's three per-seed secondary contrasts beneath it), Table~\ref{tab:five} each headline recipe's five trajectories and Table~\ref{tab:e34fmt} the format, stop
and cap rates. Two descriptive readouts carry no label and no interval. For Falcon-E, the best deployed strict accuracy
among the five in-run endpoints minus the best online one is $+0.20$ pp (63.28\% at $6.25\times10^{-6}$ minus 63.09\%
on the second $3.125\times10^{-6}$ trajectory; the same under last-number); it is not the maximum over the entire
displayed rate grid and not a fixed-seed learning-rate frontier, and selection and evaluation use the same development
rows. For BitCPM, a cross-run readout over two accepted earlier runs, at $3.125\times10^{-6}$ and $10^{-4}$ with one
trajectory each, gives $-7.42$ pp strict (26.56\% deployed at $10^{-4}$ minus 33.98\% online at $3.125\times10^{-6}$)
and $-7.23$ pp last-number (26.76\% minus 33.98\%); it is descriptive only, and no between-rate test is made.
\begin{table}[H]
\centering\scriptsize
\setlength{\tabcolsep}{3pt}
\caption{All 23 primary contrasts of the new-seed, rate and reserve-row study (\S\ref{sec:newevidence}), every one evaluable: Parts A and B's documented-path contrasts, A-BN and A-long here; Part B's six pooled remedy rows, each with its three per-seed secondary contrasts beneath it, and Part C's reserve rows in the directly adjacent Table~\ref{tab:e34rem}. Deployed (documented, fixed or projected) minus online greedy GSM8K accuracy in points, with 95\% paired row-bootstrap intervals (20,000 draws). Part A: Falcon-E rates; Part B: three new seeds per headline recipe with both remedies, the 2B-4T rate comparison (A-BN) and a longer Falcon-E recipe (A-long); Part C: the 607 reserve rows. $^*$A-BN's three strict intervals are Bonferroni-simultaneous; every other interval, and every last-number interval, is marginal. Every non-inferiority result uses $\delta = 4$ pp, the inherited margin: $\delta = 4$ pp is the inherited resolution-based margin from the 512-row dev export contrasts (median 95\% half-width, rounded down); it was not re-derived for the 607 reserve rows. The plain-language results map to the registry's labels in Table~\ref{tab:key}.}
\label{tab:e34}
\begin{adjustbox}{max width=\textwidth}
\begin{tabular}{llllll}
\toprule
part & comparison & rows & $\Delta$ strict & $\Delta$ last-number & result \\
\midrule
A & Falcon-E, $3.125\times10^{-6}$, second trajectory & dev & \iv{$-62.89$}{$-67.19$, $-58.59$} & \iv{$-11.72$}{$-16.80$, $-6.64$} & export lowers accuracy; last-number also lower \\
A & Falcon-E, $6.25\times10^{-6}$ (lowest online dev loss) & dev & \iv{$+6.25$}{$+1.76$, $+10.94$} & \iv{$+6.25$}{$+1.76$, $+10.94$} & non-inferior at the tuned rate, detectably above \\
A & Falcon-E, export at $2\times10^{-5}$ (lowest exported dev loss) $-$ online at $3.125\times10^{-6}$ & dev & \iv{$-8.98$}{$-13.67$, $-4.49$} & \iv{$-9.18$}{$-13.87$, $-4.69$} & lower; more than 4 points below; last-number also lower \\
\midrule
B & Falcon-E-1B-Base, new seed 1 & dev & \iv{$-60.16$}{$-64.26$, $-55.86$} & \iv{$-13.28$}{$-18.36$, $-8.01$} & lower at all three seeds; last-number also lower \\
 & Falcon-E-1B-Base, new seed 2 & dev & \iv{$-61.13$}{$-65.43$, $-56.84$} & \iv{$-16.60$}{$-21.88$, $-11.52$} & last-number also lower \\
 & Falcon-E-1B-Base, new seed 3 & dev & \iv{$-58.59$}{$-62.89$, $-54.30$} & \iv{$-8.59$}{$-13.87$, $-3.32$} & last-number also lower \\
B & BitNet 2B-4T, new seed 1 & dev & \iv{$-23.83$}{$-29.10$, $-18.55$} & \iv{$+4.49$}{$-0.20$, $+9.18$} & lower at all three seeds; last-number unresolved \\
 & BitNet 2B-4T, new seed 2 & dev & \iv{$-29.69$}{$-34.77$, $-24.61$} & \iv{$+8.98$}{$+4.30$, $+13.67$} & last-number higher \\
 & BitNet 2B-4T, new seed 3 & dev & \iv{$-25.00$}{$-30.47$, $-19.53$} & \iv{$+7.42$}{$+2.54$, $+12.50$} & last-number higher \\
B & BitCPM-CANN-0.5B, new seed 1 & dev & \iv{$-33.20$}{$-37.50$, $-29.10$} & \iv{$-21.09$}{$-25.20$, $-16.99$} & lower at all three seeds; last-number also lower \\
 & BitCPM-CANN-0.5B, new seed 2 & dev & \iv{$-35.74$}{$-39.84$, $-31.64$} & \iv{$-25.20$}{$-29.49$, $-20.90$} & last-number also lower \\
 & BitCPM-CANN-0.5B, new seed 3 & dev & \iv{$-32.23$}{$-36.33$, $-28.32$} & \iv{$-19.92$}{$-24.22$, $-15.62$} & last-number also lower \\
A-BN & BitNet 2B-4T, bf16 save at $3\times10^{-5}$ & dev & \iv{$-64.84$}{$-69.92$, $-59.77$}$^*$ & \iv{$-5.08$}{$-9.96$, $-0.20$} & lower \\
 & BitNet 2B-4T, bf16 save at $10^{-4}$ (lowest online dev loss) & dev & \iv{$-27.54$}{$-34.18$, $-20.90$}$^*$ & \iv{$+7.62$}{$+2.93$, $+12.30$} & lower; the label's rate \\
 & BitNet 2B-4T, bf16 save at $3\times10^{-4}$ & dev & \iv{$+8.98$}{$+4.49$, $+13.67$}$^*$ & \iv{$+9.18$}{$+5.47$, $+12.89$} & higher \\
A-long & Falcon-E, 1,200 steps at $3.125\times10^{-6}$ & dev & \iv{$+1.95$}{$-2.34$, $+6.25$} & \iv{$+2.15$}{$-2.15$, $+6.45$} & no detectable difference \\
\bottomrule
\end{tabular}
\end{adjustbox}
\end{table}

\begin{table}[H]
\centering\scriptsize
\setlength{\tabcolsep}{3pt}
\caption{Continuation of Table~\ref{tab:e34}, with its conventions: Part B's six pooled remedy rows (primaries), each with its three per-seed secondary contrasts indented beneath it, then Part C's nine reserve-row contrasts. Pooled rows average each row over the three fixed new trajectories and are declared on strict accuracy only; the per-seed rows are secondary and carry no primary label. Every non-inferiority result uses $\delta = 4$ pp, the inherited margin: $\delta = 4$ pp is the inherited resolution-based margin from the 512-row dev export contrasts (median 95\% half-width, rounded down); it was not re-derived for the 607 reserve rows. $^\dagger$The boundary flag of the four marked primary records: an interval endpoint within 0.4 pp of a threshold; all four approach zero (\S\ref{sec:newevidence}). $^\ddagger$The same 0.4 pp boundary field of a per-seed secondary record, displayed as recorded: three approach zero, and Falcon-E's seed-2 projection approaches the $-4$ pp non-inferiority margin. A marker discloses this diagnostic; it changes no label, margin or interval.}
\label{tab:e34rem}
\begin{adjustbox}{max width=\textwidth}
\begin{tabular}{llllll}
\toprule
part & comparison & rows & $\Delta$ strict & $\Delta$ last-number & result \\
\midrule
B & Falcon-E-1B-Base, fixed export, pooled over the three new seeds & dev & \iv{$+0.20$}{$-1.11$, $+1.50$} & (strict only) & pooled non-inferior \\
 & \quad seed 1 (secondary) & dev & \iv{$+0.20$}{$-2.15$, $+2.54$} & \iv{$+0.20$}{$-2.15$, $+2.54$} & non-inferior (secondary) \\
 & \quad seed 2 (secondary) & dev & \iv{$-0.78$}{$-3.32$, $+1.76$} & \iv{$-0.78$}{$-3.32$, $+1.76$} & non-inferior (secondary) \\
 & \quad seed 3 (secondary) & dev & \iv{$+1.17$}{$-1.17$, $+3.52$} & \iv{$+1.37$}{$-0.98$, $+3.52$} & non-inferior (secondary) \\
 & Falcon-E-1B-Base, projected export, pooled over the three new seeds & dev & \iv{$+0.07$}{$-1.24$, $+1.43$} & (strict only) & pooled non-inferior \\
 & \quad seed 1 (secondary) & dev & \iv{$-0.78$}{$-3.12$, $+1.56$} & \iv{$-0.78$}{$-3.12$, $+1.56$} & non-inferior (secondary) \\
 & \quad seed 2 (secondary) & dev & \iv{$-1.56$}{$-3.91$, $+0.78$}$^\ddagger$ & \iv{$-1.56$}{$-3.91$, $+0.78$} & non-inferior (secondary) \\
 & \quad seed 3 (secondary) & dev & \iv{$+2.54$}{$+0.59$, $+4.69$} & \iv{$+2.54$}{$+0.59$, $+4.69$} & non-inferior, detectably above (secondary) \\
 & BitNet 2B-4T, fixed export, pooled over the three new seeds & dev & \iv{$+1.30$}{$+0.20$, $+2.41$}$^\dagger$ & (strict only) & pooled non-inferior, detectably above \\
 & \quad seed 1 (secondary) & dev & \iv{$+1.95$}{$-0.20$, $+4.10$}$^\ddagger$ & \iv{$+1.95$}{$-0.20$, $+4.10$} & non-inferior (secondary) \\
 & \quad seed 2 (secondary) & dev & \iv{$+0.98$}{$-0.78$, $+2.73$} & \iv{$+0.98$}{$-0.78$, $+2.73$} & non-inferior (secondary) \\
 & \quad seed 3 (secondary) & dev & \iv{$+0.98$}{$-0.78$, $+2.93$} & \iv{$+0.98$}{$-0.78$, $+2.93$} & non-inferior (secondary) \\
 & BitNet 2B-4T, projected export, pooled over the three new seeds & dev & \iv{$+1.17$}{$0.00$, $+2.35$}$^\dagger$ & (strict only) & pooled non-inferior \\
 & \quad seed 1 (secondary) & dev & \iv{$+1.17$}{$-0.98$, $+3.32$} & \iv{$+1.17$}{$-0.98$, $+3.32$} & non-inferior (secondary) \\
 & \quad seed 2 (secondary) & dev & \iv{$+0.59$}{$-1.37$, $+2.54$} & \iv{$+0.59$}{$-1.37$, $+2.54$} & non-inferior (secondary) \\
 & \quad seed 3 (secondary) & dev & \iv{$+1.76$}{$-0.59$, $+4.10$} & \iv{$+1.76$}{$-0.59$, $+4.10$} & non-inferior (secondary) \\
 & BitCPM-CANN-0.5B, fixed export, pooled over the three new seeds & dev & \iv{$+0.39$}{$-0.52$, $+1.30$} & (strict only) & pooled non-inferior \\
 & \quad seed 1 (secondary) & dev & \iv{$+0.98$}{$-0.39$, $+2.34$}$^\ddagger$ & \iv{$+0.98$}{$-0.39$, $+2.34$} & non-inferior (secondary) \\
 & \quad seed 2 (secondary) & dev & \iv{$+0.20$}{$-1.17$, $+1.56$} & \iv{$+0.20$}{$-1.17$, $+1.56$} & non-inferior (secondary) \\
 & \quad seed 3 (secondary) & dev & \iv{$0.00$}{$-1.76$, $+1.76$} & \iv{$0.00$}{$-1.76$, $+1.76$} & non-inferior (secondary) \\
 & BitCPM-CANN-0.5B, projected export, pooled over the three new seeds & dev & \iv{$+0.72$}{$-0.20$, $+1.63$}$^\dagger$ & (strict only) & pooled non-inferior \\
 & \quad seed 1 (secondary) & dev & \iv{$+0.20$}{$-1.17$, $+1.56$} & \iv{$+0.20$}{$-1.17$, $+1.56$} & non-inferior (secondary) \\
 & \quad seed 2 (secondary) & dev & \iv{$+1.37$}{$+0.20$, $+2.73$}$^\ddagger$ & \iv{$+1.37$}{$+0.20$, $+2.73$} & non-inferior, detectably above (secondary) \\
 & \quad seed 3 (secondary) & dev & \iv{$+0.59$}{$-1.17$, $+2.34$} & \iv{$+0.59$}{$-1.17$, $+2.34$} & non-inferior (secondary) \\
\midrule
C & Falcon-E-1B-Base, documented $-$ online & reserve & \iv{$-53.54$}{$-57.50$, $-49.59$} & \iv{$-12.52$}{$-17.46$, $-7.58$} & same direction as dev; last-number also lower \\
 & BitNet 2B-4T, documented $-$ online & reserve & \iv{$-27.51$}{$-32.29$, $-22.73$} & \iv{$+3.95$}{$-0.33$, $+8.24$} & same direction as dev; last-number unresolved \\
 & BitCPM-CANN-0.5B, documented $-$ online & reserve & \iv{$-31.80$}{$-35.58$, $-28.01$} & \iv{$-19.77$}{$-24.05$, $-15.49$} & same direction as dev; last-number also lower \\
 & Falcon-E-1B-Base, fixed $-$ online & reserve & \iv{$+1.15$}{$-1.15$, $+3.46$} & \iv{$+0.99$}{$-1.32$, $+3.29$} & non-inferior \\
 & Falcon-E-1B-Base, projected $-$ online & reserve & \iv{$-0.33$}{$-2.47$, $+1.81$} & \iv{$-0.49$}{$-2.64$, $+1.65$} & non-inferior \\
 & BitNet 2B-4T, fixed $-$ online & reserve & \iv{$-1.32$}{$-3.13$, $+0.33$}$^\dagger$ & \iv{$-1.32$}{$-3.13$, $+0.33$} & non-inferior \\
 & BitNet 2B-4T, projected $-$ online & reserve & \iv{$-1.15$}{$-3.13$, $+0.82$} & \iv{$-1.15$}{$-3.13$, $+0.82$} & non-inferior \\
 & BitCPM-CANN-0.5B, fixed $-$ online & reserve & \iv{$+0.16$}{$-1.15$, $+1.48$} & \iv{$+0.16$}{$-1.15$, $+1.48$} & non-inferior \\
 & BitCPM-CANN-0.5B, projected $-$ online & reserve & \iv{$-0.49$}{$-1.81$, $+0.82$} & \iv{$-0.49$}{$-1.81$, $+0.82$} & non-inferior \\
\bottomrule
\end{tabular}
\end{adjustbox}
\end{table}

\begin{table}[H]
\centering\scriptsize
\setlength{\tabcolsep}{3pt}
\caption{Each headline recipe's five trajectories, listed side by side and not summarized as a seed-variance estimate: greedy accuracy on the 512 dev rows, strict / last-number, online and after the documented path, and documented minus online in points. The two earlier trajectories are the accepted records of \S\ref{sec:endpoints} and Part A's second Falcon-E trajectory.}
\label{tab:five}
\begin{adjustbox}{max width=\textwidth}
\begin{tabular}{lllllll}
\toprule
model & trajectory & seed & source & online & documented & documented $-$ online \\
\midrule
Falcon-E-1B-Base & A1 & 2026091700 & earlier run & 58.79\% / 58.79\% & 0.78\% / 47.07\% & $-58.01$ / $-11.72$ \\
 & FE-B1 & 2026092301 & earlier run & 63.09\% / 63.09\% & 0.20\% / 51.37\% & $-62.89$ / $-11.72$ \\
 & FE-S1 & 2026101401 & new seed & 60.55\% / 60.55\% & 0.39\% / 47.27\% & $-60.16$ / $-13.28$ \\
 & FE-S2 & 2026101402 & new seed & 61.52\% / 61.52\% & 0.39\% / 44.92\% & $-61.13$ / $-16.60$ \\
 & FE-S3 & 2026101403 & new seed & 58.98\% / 58.98\% & 0.39\% / 50.39\% & $-58.59$ / $-8.59$ \\
\addlinespace
BitNet 2B-4T & C1 & 2026091700 & earlier run & 58.40\% / 58.20\% & 30.86\% / 65.82\% & $-27.54$ / $+7.62$ \\
 & C2 & 2026092301 & earlier run & 58.98\% / 58.98\% & 35.35\% / 66.60\% & $-23.63$ / $+7.62$ \\
 & BN-S1 & 2026101401 & new seed & 59.38\% / 59.38\% & 35.55\% / 63.87\% & $-23.83$ / $+4.49$ \\
 & BN-S2 & 2026101402 & new seed & 58.20\% / 58.20\% & 28.52\% / 67.19\% & $-29.69$ / $+8.98$ \\
 & BN-S3 & 2026101403 & new seed & 58.40\% / 58.40\% & 33.40\% / 65.82\% & $-25.00$ / $+7.42$ \\
\addlinespace
BitCPM-CANN-0.5B & CPM-B1 & 2026091700 & earlier run & 33.98\% / 33.98\% & 0.98\% / 12.11\% & $-33.01$ / $-21.88$ \\
 & CPM-B2 & 2026092301 & earlier run & 36.13\% / 36.13\% & 0.39\% / 13.09\% & $-35.74$ / $-23.05$ \\
 & CPM-S1 & 2026101401 & new seed & 34.18\% / 34.18\% & 0.98\% / 13.09\% & $-33.20$ / $-21.09$ \\
 & CPM-S2 & 2026101402 & new seed & 36.52\% / 36.52\% & 0.78\% / 11.33\% & $-35.74$ / $-25.20$ \\
 & CPM-S3 & 2026101403 & new seed & 33.01\% / 33.01\% & 0.78\% / 13.09\% & $-32.23$ / $-19.92$ \\
\bottomrule
\end{tabular}
\end{adjustbox}
\end{table}

\begin{table}[H]
\centering\scriptsize
\setlength{\tabcolsep}{3pt}
\caption{Format-match, EOS-stop and cap-hit rates, online $\to$ documented, at the three new seeds and on the 607 reserve rows. They describe the generations; no causal allocation between format and ability is made.}
\label{tab:e34fmt}
\begin{tabular}{lllll}
\toprule
model & rows & format match & EOS stop & cap hit \\
\midrule
Falcon-E-1B-Base & new seed 1 (dev) & 96.9\% $\to$ 17.0\% & 96.9\% $\to$ 66.4\% & 3.1\% $\to$ 33.6\% \\
 & new seed 2 (dev) & 96.3\% $\to$ 21.3\% & 96.3\% $\to$ 59.8\% & 3.7\% $\to$ 40.2\% \\
 & new seed 3 (dev) & 96.9\% $\to$ 14.8\% & 96.9\% $\to$ 71.1\% & 3.1\% $\to$ 28.9\% \\
 & reserve rows & 98.4\% $\to$ 22.4\% & 98.4\% $\to$ 60.5\% & 1.6\% $\to$ 39.5\% \\
\addlinespace
BitNet 2B-4T & new seed 1 (dev) & 98.0\% $\to$ 48.4\% & 98.0\% $\to$ 11.1\% & 2.0\% $\to$ 88.9\% \\
 & new seed 2 (dev) & 98.8\% $\to$ 35.7\% & 98.8\% $\to$ 16.0\% & 1.2\% $\to$ 84.0\% \\
 & new seed 3 (dev) & 98.4\% $\to$ 41.8\% & 98.4\% $\to$ 8.6\% & 1.6\% $\to$ 91.4\% \\
 & reserve rows & 98.5\% $\to$ 38.9\% & 98.5\% $\to$ 8.1\% & 1.5\% $\to$ 91.9\% \\
\addlinespace
BitCPM-CANN-0.5B & new seed 1 (dev) & 96.1\% $\to$ 3.3\% & 96.1\% $\to$ 52.3\% & 3.9\% $\to$ 47.7\% \\
 & new seed 2 (dev) & 95.3\% $\to$ 2.5\% & 95.3\% $\to$ 49.4\% & 4.7\% $\to$ 50.6\% \\
 & new seed 3 (dev) & 96.9\% $\to$ 2.5\% & 96.7\% $\to$ 52.1\% & 3.3\% $\to$ 47.9\% \\
 & reserve rows & 96.5\% $\to$ 3.6\% & 96.4\% $\to$ 50.7\% & 3.6\% $\to$ 49.3\% \\
\bottomrule
\end{tabular}
\end{table}

\textbf{The other-task and training-forward study} (\S\ref{sec:endpoints}). Table~\ref{tab:e35} lists its 11 primary
contrasts, all evaluable, with the three online-minus-base qualifiers, and Table~\ref{tab:e35states} each state's
accuracy. Falcon-E's BoolQ arm trains on the question-weighted 9,385-row pool (several questions per passage), whose
4,800-row stream draws 4,800 distinct rows and leaves 4,585 undrawn (0.51 passes); the ARC arm's stream covers its
3,358 rows (1.43 passes) and 2B-4T's BoolQ arm its 4,096 (1.17 passes). Validation cross-entropy on the rosters was
recorded at steps 0, 150 and 300 (2B-4T: 0 and 300) as a technical anchor; it selected no checkpoint, duration, task,
rate or retry. Rank choice reads the candidates' float32 log-probabilities at the final prompt position, with the body
in each state's declared context; this float32 head was adopted after a training-row build smoke and before any
roster outcome. Build checks detected context-dependent score and decision differences, reported separately, and the
bootstrap intervals do not quantify sensitivity to other numerical execution contexts. ARC's Easy and Challenge
breakdowns are descriptive only. Part DA's ordinary comparator is the fp32 masters of Falcon-E's A1 endpoint as
retrained for the remedy studies (\S\ref{sec:remedies}), evaluated afresh in this run.
\begin{table}[H]
\centering\scriptsize
\setlength{\tabcolsep}{3pt}
\caption{The other-task and training-forward study (\S\ref{sec:endpoints}): all 11 primary contrasts, every one evaluable, and the three online-minus-base qualifiers, which are never counted. Part T: rank-choice accuracy on 512 validation rows per case, by fixed prompt-only, batch-one scoring with a float32 output head and exact ties scored incorrect; one trajectory per case. Part DA: greedy GSM8K dev accuracy, strict and last-number; one trajectory per procedure. Differences in points with 95\% paired row-bootstrap intervals (20,000 draws), marginal per contrast; there is no joint coverage claim and no package-level success claim. Part T's non-inferiority uses $\delta = 4$ pp, inherited from the GSM8K dev export contrasts (median 95\% half-width, rounded down); it was not derived or validated for ARC or BoolQ rank-choice accuracy. Part DA's uses the inherited resolution-based margin from the 512-row dev export contrasts (median 95\% half-width, rounded down). The retained-gain ratio is (documented $-$ base)/(online $-$ base), unclipped, reported where the gain was established; it is a ratio to each case's own measured gain. $^\dagger$The boundary flag: an interval endpoint within 0.4 pp of a threshold; all three approach zero, not the $-4$ pp margin. The plain-language results map to the registry's labels in Table~\ref{tab:key}.}
\label{tab:e35}
\begin{adjustbox}{max width=\textwidth}
\begin{tabular}{llllll}
\toprule
part & case & comparison & $\Delta$ (rank choice; DA: strict) & $\Delta$ last-number & result \\
\midrule
T & Falcon-E-1B-Base, ARC (validation mixture) & online $-$ base (qualifier) & \iv{$+10.16$}{$+6.45$, $+13.87$} & -- & fine-tuning gain established (qualifier) \\
 &  & documented $-$ online & \iv{$-2.93$}{$-6.25$, $+0.39$}$^\dagger$ & -- & no detectable difference; inconclusive at $\delta = 4$ pp; retained-gain ratio 0.712 \\
 &  & fixed $-$ online & \iv{$0.00$}{$-0.78$, $+0.78$} & -- & non-inferior \\
 &  & projected $-$ online & \iv{$-0.39$}{$-1.37$, $+0.59$} & -- & non-inferior \\
\midrule
T & Falcon-E-1B-Base, BoolQ & online $-$ base (qualifier) & \iv{$+25.39$}{$+20.12$, $+30.47$} & -- & fine-tuning gain established (qualifier) \\
 &  & documented $-$ online & \iv{$-13.48$}{$-17.58$, $-9.38$} & -- & export lowers accuracy; more than 4 points below; retained-gain ratio 0.469 \\
 &  & fixed $-$ online & \iv{$0.00$}{$-0.78$, $+0.78$} & -- & non-inferior \\
 &  & projected $-$ online & \iv{$+0.20$}{$-0.59$, $+0.98$} & -- & non-inferior \\
\midrule
T & BitNet 2B-4T, BoolQ & online $-$ base (qualifier) & \iv{$+17.97$}{$+13.67$, $+22.07$} & -- & fine-tuning gain established (qualifier) \\
 &  & documented $-$ online & \iv{$-4.30$}{$-7.42$, $-1.37$} & -- & export lowers accuracy; inconclusive at $\delta = 4$ pp; retained-gain ratio 0.761 \\
 &  & fixed $-$ online & \iv{$+0.20$}{$0.00$, $+0.59$}$^\dagger$ & -- & non-inferior \\
 &  & projected $-$ online & \iv{$+0.20$}{$-0.39$, $+0.98$}$^\dagger$ & -- & non-inferior \\
\midrule
DA & Falcon-E-1B-Base, GSM8K dev (A1's recipe) & export-arithmetic training's real-tool export $-$ ordinary training, online & \iv{$+2.93$}{$-1.17$, $+7.03$} & \iv{$+2.93$}{$-1.17$, $+7.03$} & non-inferior; no detectable difference under either rule \\
 &  & export-arithmetic training's real-tool export $-$ ordinary training's real-tool export & \iv{$+60.94$}{$+56.64$, $+65.04$} & \iv{$+14.65$}{$+9.77$, $+19.73$} & higher under both rules \\
\bottomrule
\end{tabular}
\end{adjustbox}
\end{table}

\begin{table}[H]
\centering\scriptsize
\setlength{\tabcolsep}{4pt}
\caption{The other-task and training-forward study's states. Top: Part T's rank-choice accuracy (\%) per state, with the number of rows whose top-two score margin is below 0.5 in parentheses; no exact tie and no non-finite score occurred in any state. The 0.5 count is a prespecified, non-calibrated small-margin diagnostic, not an error bound, and no row is dropped, reweighted or rescored because of it. 2B-4T's BoolQ arm was trained on the answer token only, Falcon-E's arms on the answer and EOS. Bottom: Part DA's greedy GSM8K dev accuracy (\%), strict / last-number. Between the codes and scales captured from the export-arithmetic training's own forward and the real exporter's output, 0 code and 0 scale mismatches were counted over 1,585,446,912 codes in 168 projections; the non-projection casting and the inference path still differ.}
\label{tab:e35states}
\begin{tabular}{lrrrrr}
\toprule
case & base & online & documented & fixed & projected \\
\midrule
Falcon-E-1B-Base, ARC (validation mixture) & 72.85 (126) & 83.01 (16) & 80.08 (81) & 83.01 (16) & 82.62 (23) \\
Falcon-E-1B-Base, BoolQ & 58.01 (297) & 83.40 (68) & 69.92 (166) & 83.40 (65) & 83.59 (67) \\
BitNet 2B-4T, BoolQ & 68.36 (40) & 86.33 (38) & 82.03 (54) & 86.52 (34) & 86.52 (39) \\
\bottomrule
\end{tabular}

\medskip
\begin{tabular}{lr}
\toprule
state & strict / last-number \\
\midrule
ordinary training, online (the fresh comparator) & 58.79 / 58.79 \\
ordinary training, real-tool export & 0.78 / 47.07 \\
export-arithmetic training, online (its own training forward) & 60.55 / 60.74 \\
export-arithmetic training, real-tool export & 61.72 / 61.72 \\
base, ordinary forward & 0.20 / 23.83 \\
base, export-arithmetic forward & 0.20 / 25.00 \\
\bottomrule
\end{tabular}
\end{table}

\section{The on-grid bf16 master-weight study in detail}
\label{app:training}
\begin{table}[H]
\centering\scriptsize
\setlength{\tabcolsep}{4pt}
\caption{The on-grid bf16 master-weight study: 300-step GSM8K fine-tuning with bf16-grid latent writes. Strict accuracy (\%) of the deployed target; $\Delta$, in points, is that minus the freshly evaluated fp32 online state, with the family-adjusted [1.25, 98.75] interval for the emulations and the central 95\% interval for native bf16. Codes: evaluation-context codes at the endpoint that differ from step 0. Under the declared rules, native bf16 left 2B-4T's endpoint codes equal to those at step 0 (frozen) and changed one Falcon-E code (not frozen), and all four emulations were non-inferior and met the method-success rule (Table~\ref{tab:key}).}
\label{tab:training}
\begin{adjustbox}{max width=\columnwidth}
\begin{tabular}{@{}lrlr@{}}
\toprule
writer & strict & $\Delta$ [interval] & codes \\
\midrule
\multicolumn{4}{@{}l}{\emph{BitNet 2B-4T} (target: bf16 save and reload; fp32 58.40\%)} \\
native bf16 & 57.62 & $-0.78$ [$-5.08$, $+3.52$] & 0 \\
SR emulation & 59.18 & $+0.78$ [$-3.52$, $+5.08$] & 29{,}814{,}178 \\
KH emulation & 59.38 & $+0.98$ [$-3.91$, $+5.86$] & 2{,}822{,}197 \\
\addlinespace
\multicolumn{4}{@{}l}{\emph{Falcon-E-1B-Base} (target: onebitllms export; fp32 58.79\%)} \\
native bf16 & 0.20 & $-58.59$ [$-62.89$, $-54.30$] & 1 \\
SR emulation & 60.35 & $+1.56$ [$-2.73$, $+6.05$] & 16{,}648{,}041 \\
KH emulation & 59.57 & $+0.78$ [$-3.71$, $+5.47$] & 4{,}380{,}868 \\
\bottomrule
\end{tabular}
\end{adjustbox}
\end{table}

\textbf{Through the save.} An on-grid writer makes the bf16 save an identity on the latents by construction: the cast
left latents and codes unchanged in every arm, native included, so this is a design property rather than evidence for
SR or KH. It is the counterpart of \S\ref{sec:endpoints}, where fp32 training changed codes that the bf16 save then
reverted. Falcon-E's documented export still changed 26,319,163 (SR) and 27,763,368 (KH) of its 1,585,446,912 codes
through its bf16 arithmetic, reverting 3,963,759 and 1,298,202 of the training's code additions, and the exported
states still met the criterion.

\textbf{Further native and emulated results} (\S\ref{sec:training}). NB and the emulations also differ in gradient
storage and accumulation, the optimizer kernel, the clipping-norm dtype and the forward amplitude dtype, so
comparisons between them compare procedures. In the native 2B-4T run, no training-context code
transition occurred over steps 1--299; the final training-context transition was not observed. A run that trained
only the non-projection parameters, whose clipping norm therefore covered only those gradients, matched the native run
in cross-entropy within $\pm0.01$ nats, while last-number accuracy was inconclusive at $\pm2$ points, so the two are
not jointly equivalent. Against native bf16, Falcon-E's SR and KH targets show descriptive strict-accuracy improvements
of about 60 points. In 2B-4T there is no established strict-accuracy improvement over native bf16, while cross-entropy
is lower by 0.108 (SR) and 0.085 (KH) nats per token.

\textbf{Learning rate} (native 2B-4T). At $10^{-3}$ no training-context code transition occurred over steps 1--299, and
strict accuracy was 46.88\%; at $4\times10^{-3}$ codes changed and strict accuracy was 5.08\%. A declared one-sided
forecast---no fixed-scale latent-induced code events in tensors whose half-ULP at the threshold is at least 1.1 times
the peak learning rate---held at all three rates, with every scoped tensor-step covered. It does not predict which
tensors move, and we draw no causal link between code movement and accuracy.

\textbf{Scope.} Two models, one task (GSM8K dev, 512 rows), one recipe per model with one learning rate per model plus
two additional BitNet onset learning rates (1e-3, 4e-3), PagedAdamW8bit, the declared harness and states; one training
trajectory and one SR seed per arm. The dev rows were used before and informed the design, the intervals do not measure seed variability, the 4-point margin is
\S\ref{sec:remedies}'s inherited resolution-based margin, and the family coverage is nominal under the bootstrap
approximation.

\textbf{Protocol.} The study was pre-registered with frozen analysis code and reporting rules, a manifest of
source, record and library hashes, and a GPU build check on smoke runs that passed before launch. Three successor packages
fixed technical defects in the controller and the analyzer's input checks before launch, and a fourth, during the run,
only raised the GPU reservation by owner decision (below); the training and evaluation workers stayed byte-identical to
the frozen originals, and no label rule changed. An AI
pre-launch review accepted the package with conditions, and AI result reviews recomputed the labels and statistics from
the saved records.

\textbf{Run history.} The free-GPU gate paused the run once, and it was relaunched when the GPU emptied. The relaunched
controller then died with the AI-assistant session that had started it, interrupting the 24th of 32 jobs, Falcon-E's
RNEe training, at step 150. Reconciling that job charged a conservative GPU-time bound, which crossed the declared
12 GPU-hour reservation; under the declared stop-and-ask rule the owner raised the reservation to 26 GPU-hours, in a
successor package that changed nothing else. The interrupted training was retried from the start and reproduced all
303 of the interrupted run's step-log records exactly. 23 outcomes were produced under the earlier package and 9 under
the successor. Other processes shared the GPU during parts of three jobs; the controller checked the GPU before each
job, not continuously, so the job times are not an exclusive-GPU benchmark. A check specified to precede each
successor's build was recorded for the first successor only, and for the last one after the run.

\textbf{Measurement.} Codes are read from the installed training quantizer's own calls in two compiled contexts: the
training forward (states 0--299) and the evaluation forward (step 0 and the endpoint). The freeze label compares the
evaluation-context endpoint with step 0; the training-context labels in Table~\ref{tab:app33a} cover states 0--299.
The onset forecast calls a 2B-4T tensor a forecast mover when 1.1 times the peak learning rate exceeds half the bf16 ULP
of its threshold, and tests only that forecast non-movers show no fixed-scale, latent-induced code events, using an
eager decomposition of each step's code changes into latent- and scale-induced parts; all 63{,}000, 61{,}200 and
2{,}100 scoped tensor-steps were covered. KH's compensation stayed within half a ULP at the endpoint in all 332 and 219
parameter tensors. The frozen logger records neither native realized-write histograms nor a base-to-endpoint eager
channel table, so neither is reported.

\textbf{Public fine-tunes.} Within a declared 644-repository search universe, nine analysed full checkpoints passed
our metadata/dtype/finite-value filter. Six have zero ternary projection-code differences from the bf16 BitNet 2B-4T
release under the audit quantizer; the other three differ at 20{,}934{,}139, 621{,}899 and 18{,}337 coordinates. Every
listed repository has a recorded disposition, and every accessible candidate required by the declared screening rules
was analysed. Fifteen listed repositories were gated and remain unassessed; two advertise BitNetForCausalLM with
bf16-sized weight files.
\begin{itemize}
\item \emph{Universe.} It combines the 2B-4T base-model tree listed on 2026-10-09, a seven-term Hugging Face search
listed on 2026-10-13 and an earlier scout's lists. It is a union of dated listings, not an exhaustive inventory:
private, later-created and otherwise undiscovered repositories are outside it, and the gated members are inside it but
unassessed.
\item \emph{Filter.} The unchanged filter was written after the earlier counts were known and before the two newly
eligible checkpoints were analysed. The original training procedures may be unknown, and a full checkpoint does not
prove full-parameter training. The census is neither a prevalence estimate nor evidence that community training used
bf16 parameters.
\item \emph{The two newly eligible checkpoints}, from one uploader and so not independent training replications,
store ternary-valued projections (at most one nonzero magnitude per tensor) rather than continuous latents. One of them
has zero code differences while nearly all of its stored projection values differ from the release's, so zero code
differences show neither checkpoint identity, nor an unchanged forward, nor failed learning. The earlier seven gave five
with zero differences and two with nonzero differences; the added two give one of each.
\item \emph{Identity.} Byte identity with the analysed files is recorded for the 22 checkpoints analysed on 2026-10-13
and for two earlier repositories only, and equal codes do not establish equal scales, non-projection parameters, function or task
performance.
\end{itemize}
\begin{table}[H]
\centering\scriptsize
\setlength{\tabcolsep}{3pt}
\caption{The on-grid bf16 master-weight study's training-context code record (installed codes, grad prefix: states 0--299, the final training-context transition not observed): cumulative transitions and net differences from the base at state 299 (additions, removals; no sign reversals occurred), return events, and the one-sided onset forecast for the native 2B-4T arms. P-BN trains only the non-projection parameters. Learning rates: $10^{-4}$ (2B-4T), $3.125\times10^{-6}$ (Falcon-E); NB-BN-1e-3 and NB-BN-4e-3 at $10^{-3}$ and $4\times10^{-3}$.}
\label{tab:app33a}
\begin{tabular}{llllrl}
\toprule
arm & grad-prefix label & cumulative (add, rem) & net at 299 (add, rem) & returns & onset \\
\midrule
NB-BN & frozen & 0, 0 & 0, 0 & 0 & holds \\
RNEe-BN & frozen & 0, 0 & 0, 0 & 0 & -- \\
SR-BN & not frozen & 17{,}721{,}440, 17{,}732{,}976 & 14{,}901{,}321, 14{,}912{,}857 & 2{,}820{,}119 & -- \\
KH-BN & not frozen & 4{,}034{,}663, 4{,}149{,}112 & 1{,}353{,}874, 1{,}468{,}323 & 2{,}680{,}789 & -- \\
P-BN & frozen & 0, 0 & 0, 0 & 0 & -- \\
NB-BN-1e-3 & frozen & 0, 0 & 0, 0 & 0 & holds \\
NB-BN-4e-3 & not frozen & 11{,}995{,}270, 11{,}901{,}877 & 9{,}196{,}344, 9{,}102{,}951 & 2{,}798{,}926 & holds \\
NB-FE & not frozen & 1, 0 & 1, 0 & 0 & -- \\
RNEe-FE & frozen & 0, 0 & 0, 0 & 0 & -- \\
SR-FE & not frozen & 9{,}885{,}861, 9{,}680{,}564 & 8{,}426{,}669, 8{,}221{,}372 & 1{,}459{,}192 & -- \\
KH-FE & not frozen & 6{,}009{,}117, 5{,}964{,}889 & 2{,}212{,}548, 2{,}168{,}320 & 3{,}796{,}569 & -- \\
\bottomrule
\end{tabular}
\end{table}

\begin{table}[H]
\centering\scriptsize
\setlength{\tabcolsep}{3pt}
\caption{The on-grid bf16 master-weight study's descriptive paired contrasts of the deployed target states (marginal 95\% bootstrap intervals, 10,000 draws; strict and last-number in points, CE in nats per token). An improvement over a comparator needs the lower bound above 0 (CE: the upper bound below 0).}
\label{tab:app33b}
\begin{tabular}{lllll}
\toprule
target & comparator & strict & last-number & CE \\
\midrule
SR-BN & NB-BN & \iv{$+1.56$}{$-2.93$, $+6.05$} & \iv{$+1.37$}{$-3.12$, $+5.86$} & \iv{$-0.1079$}{$-0.1154$, $-0.1008$} \\
SR-BN & RNEe-BN & \iv{$+1.95$}{$-2.54$, $+6.45$} & \iv{$+1.76$}{$-2.73$, $+6.25$} & \iv{$-0.1082$}{$-0.1156$, $-0.1010$} \\
SR-BN & own online & \iv{$-0.39$}{$-2.54$, $+1.56$} & \iv{$-0.39$}{$-2.54$, $+1.56$} & \iv{$+0.0001$}{$-0.0002$, $+0.0004$} \\
KH-BN & NB-BN & \iv{$+1.76$}{$-2.73$, $+6.25$} & \iv{$+1.56$}{$-2.93$, $+6.05$} & \iv{$-0.0853$}{$-0.0917$, $-0.0793$} \\
KH-BN & RNEe-BN & \iv{$+2.15$}{$-2.34$, $+6.84$} & \iv{$+1.95$}{$-2.54$, $+6.45$} & \iv{$-0.0856$}{$-0.0919$, $-0.0795$} \\
KH-BN & own online & \iv{$-1.56$}{$-3.71$, $+0.59$} & \iv{$-1.56$}{$-3.71$, $+0.59$} & \iv{$-0.0002$}{$-0.0005$, $+0.0002$} \\
NB-BN & fresh fp32 online & \iv{$-0.78$}{$-5.08$, $+3.52$} & \iv{$-0.39$}{$-4.88$, $+4.10$} & \iv{$+0.1073$}{$+0.0998$, $+0.1151$} \\
\midrule
SR-FE & NB-FE & \iv{$+60.16$}{$+55.86$, $+64.45$} & \iv{$+36.52$}{$+31.45$, $+41.41$} & \iv{$-0.6135$}{$-0.6330$, $-0.5952$} \\
SR-FE & RNEe-FE & \iv{$+60.35$}{$+56.05$, $+64.65$} & \iv{$+36.91$}{$+31.84$, $+41.99$} & \iv{$-0.6136$}{$-0.6332$, $-0.5953$} \\
SR-FE & own online & \iv{$-0.78$}{$-4.69$, $+3.12$} & \iv{$-0.78$}{$-4.69$, $+3.12$} & \iv{$+0.0138$}{$+0.0122$, $+0.0155$} \\
KH-FE & NB-FE & \iv{$+59.38$}{$+55.08$, $+63.67$} & \iv{$+36.33$}{$+31.25$, $+41.41$} & \iv{$-0.6043$}{$-0.6238$, $-0.5862$} \\
KH-FE & RNEe-FE & \iv{$+59.57$}{$+55.27$, $+63.87$} & \iv{$+36.72$}{$+31.64$, $+41.80$} & \iv{$-0.6045$}{$-0.6239$, $-0.5862$} \\
KH-FE & own online & \iv{$+0.20$}{$-3.91$, $+4.10$} & \iv{$+0.78$}{$-3.32$, $+4.69$} & \iv{$+0.0110$}{$+0.0092$, $+0.0128$} \\
NB-FE & fresh fp32 online & \iv{$-58.59$}{$-62.89$, $-54.30$} & \iv{$-34.96$}{$-40.04$, $-29.88$} & \iv{$+0.6259$}{$+0.6073$, $+0.6459$} \\
\bottomrule
\end{tabular}
\end{table}

\section{Threshold-distance selection in full}
\label{app:selection}
\textbf{Definitions.} Each eligible coordinate $i$, within 2 bf16 ULP of the threshold and not exactly zero, gets the score
$b_i = -g_i\,a\,\Delta q_i$: the base model's first-order prediction of the loss reduction from its candidate code
change, with $g_i$ the base-model loss gradient, $a$ the tensor's amplitude and $\Delta q_i$ the change of code, so that
a positive score predicts lower loss. A level is the set of eligible coordinates in one tensor and class (additions
from code 0, removals from $\pm1$) that share one exact bf16 magnitude, and a crosser is a coordinate whose code changed
by the endpoint. Same-level lift is the crossers' share of positive scores minus that share among all eligible
coordinates of the same level, averaged over levels with crosser weights. It is a difference of shares, positive when crossers are more
often predicted to help than their level-mates.

\begin{figure}[H]
\centering
\includegraphics[width=\columnwidth]{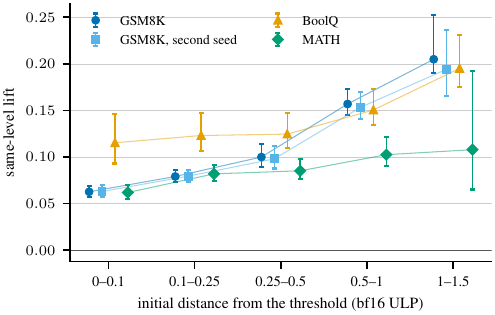}
\caption{Same-level lift (the crossers' share of positive first-order scores minus their level-mates' share) by initial
distance from the threshold on 2B-4T, at the GSM8K endpoint, its second seed, and the BoolQ and MATH endpoints. Bars are 95\% tensor-bootstrap intervals conditional on each endpoint; only bins with at
least 1,000 crossers from at least 10 tensors are drawn.}
\label{fig:lift}
\end{figure}

\textbf{The randomized intervention.} Pairs of coordinates with the same starting value were moved apart in the fp32
latent, one to half and one to one and a half times their distance, with roles swapped between two arms and the
starting computation bitwise identical (Table~\ref{tab:intervention}). The far-moved members' lift exceeds the
near-moved members' in both families. Moving a coordinate farther also makes it less likely to cross (on Falcon-E the
far-minus-near crossing probability is $-0.133$ [$-0.147$, $-0.119$]), so the result concerns what the remaining
crossings are enriched for, not a larger number of useful edits. Support depends on the outcome: strata that fail the
far-role 20-crosser floor are excluded, about 34\% of the pair weight, and on Falcon-E the supported strata lie at
smaller initial distances (pair-weighted mean 0.194 ULP, against 0.411 for the excluded ones). Falcon-E's estimate
supports the declared positive direction but falls below its frozen prediction; its supported label does not validate
that magnitude. That intervention, the Falcon-E distance intervention, was declared after an earlier observational
replication on Falcon-E had missed its frozen rule (22/25 pairs against a required 0.90), and its prediction's
estimator was chosen from that replication's data; the replication's frozen verdict, not replicated, stands. On
2B-4T, the first intervention was accepted as qualified pilot evidence, with its input bindings verified after the
verdict (Table~\ref{tab:app13}).

\begin{table}[H]
\centering\scriptsize
\setlength{\tabcolsep}{3pt}
\caption{The randomized initial-distance intervention: far-moved minus near-moved lift (95\% tensor-cluster bootstrap,
conditional on the fixed trajectories) against the frozen prediction. All three are supported under their declared
rules; the label does not validate the predicted magnitude.}
\label{tab:intervention}
\begin{adjustbox}{max width=\columnwidth}
\begin{tabular}{llrrr}
\toprule
model & $\Delta$ & prediction & tensors & excluded \\
\midrule
2B-4T, seed 1 & \iv{$+0.0372$}{$+0.0320$, $+0.0423$} & $+0.040$ & 144 & 34.5\% \\
2B-4T, seed 2 & \iv{$+0.0391$}{$+0.0330$, $+0.0449$} & $+0.040$ & 144 & 34.7\% \\
Falcon-E-1B-Base & \iv{$+0.0885$}{$+0.0746$, $+0.1023$} & $+0.125$ & 105 & 34.0\% \\
\bottomrule
\end{tabular}
\end{adjustbox}
\end{table}

\textbf{Finite edits.} In BitNet's intervention endpoints (the finite-edit reversion test, on the first 2B-4T
intervention's two arms), reverting the far-role crossers raises dev loss by $+0.00325$ [$+0.00258$, $+0.00391$] nats
per token more than reverting an equal number of near-role crossers. The far set holds all far-role crossers while the
near set is an equal-size sample, so role and coverage do not vary independently. These results support
distance-dependent selection under the joint intervention. They do not establish a benefit for individual
coordinates, a new generic threshold mechanism, or improved task performance.

\textbf{The distance contrast beyond the first endpoint} (\S\ref{sec:selection}). It is $+0.090$ [$+0.077$, $+0.108$] at
a second seed, and, under rules declared before their outcomes, $+0.035$ [$+0.003$, $+0.068$] on BoolQ and $+0.041$
[$+0.026$, $+0.060$] on MATH, in 95\% tensor-cluster bootstraps conditional on the fixed trajectories. It survives
scores computed on separate rows, and excluding the 1.53\% of coordinates whose released packed code differs from the
bf16 master.

\textbf{Further tables.} Table~\ref{tab:app12} gives the distance contrast, the same-level lift in the 0.5--1 ULP bin minus the lift below 0.1
ULP, for each endpoint and scoring. P1 counts, among tensors in which both nearest levels (the adjacent bf16 values
straddling the threshold) qualify, those whose farther nearest level has the larger lift. In Table~\ref{tab:app13},
``input bindings'' records whether the experiment's input hashes were verified after the verdict or bound before
launch, and the last row is the independent AI review's verdict (AI Usage Disclosure). Table~\ref{tab:app14}
gives the finite-edit interaction analysis's exact-partition interactions, $J = L(\text{whole}) - L(\text{quarter}) - L(\text{complement})
+ L(\text{endpoint})$, for the far set ($J_F$) and the near set ($J_N$), and their difference $D$.
\begin{table}[H]
\centering\scriptsize
\setlength{\tabcolsep}{3pt}
\caption{The distance contrast.}
\label{tab:app12}
\begin{adjustbox}{max width=\textwidth}
\begin{tabular}{lll}
\toprule
endpoint and scores & contrast [95\%] & P1 \\
\midrule
GSM8K original seed, original scores & \makecell[l]{+0.094\\{[+0.081, +0.111]}} & 41/41 \\
GSM8K original seed, separate scoring halves (256 rows) & \makecell[l]{+0.096\\{[+0.083, +0.111]}} & 41/41 \\
GSM8K second seed, original scores & \makecell[l]{+0.090\\{[+0.077, +0.108]}} & 39/40 \\
GSM8K second seed, separate scoring halves & \makecell[l]{+0.090\\{[+0.076, +0.107]}} & 40/40 \\
BoolQ (its own scores; a declared secondary) & \makecell[l]{+0.035\\{[+0.003, +0.068]}} & 29/29 \\
MATH (its own scores; a declared secondary) & \makecell[l]{+0.041\\{[+0.026, +0.060]}} & 40/41 \\
\bottomrule
\end{tabular}
\end{adjustbox}
\end{table}

\begin{table}[H]
\centering\scriptsize
\setlength{\tabcolsep}{3pt}
\caption{The intervention at two training seeds.}
\label{tab:app13}
\begin{adjustbox}{max width=\textwidth}
\begin{tabular}{lll}
\toprule
 & first 2B-4T intervention (seed 2026091700) & its seed-2 repeat (seed 2026092301) \\
\midrule
arms & A, B, C (C untouched) & A, B; historical C as descriptive context \\
$\Delta$ [95\%] & +0.037 [+0.032, +0.042], supported & +0.039 [+0.033, +0.045], supported \\
supported pair weight (tensors) & 65.5\% (144) & 65.3\% (144) \\
input bindings & verified after the verdict & bound before launch \\
codex & accepted as qualified pilot evidence & accepted (retrospective review of the completed run) \\
\bottomrule
\end{tabular}
\end{adjustbox}
\end{table}

\begin{table}[H]
\centering\scriptsize
\setlength{\tabcolsep}{3pt}
\caption{Exact-partition interactions (the finite-edit interaction analysis; nats per token; joint row bootstrap).}
\label{tab:app14}
\begin{adjustbox}{max width=\textwidth}
\begin{tabular}{llll}
\toprule
 & pooled & arm A & arm B \\
\midrule
J\_F (far set) & \makecell[l]{+0.00348\\{[+0.00307, +0.00388]}} & +0.00348 & +0.00349 \\
J\_N (near set) & \makecell[l]{+0.00151\\{[+0.00117, +0.00184]}} & +0.00135 & +0.00168 \\
D = J\_F $-$ J\_N & \makecell[l]{+0.00197\\{[+0.00155, +0.00240]}} & +0.00213 & +0.00181 \\
\bottomrule
\end{tabular}
\end{adjustbox}
\end{table}

\textbf{Related work on threshold distance.} Latent weights act as inertia \citep{helwegen2019latent}; latents near thresholds oscillate
\citep{nagel2022oscillations}; transition rates can be scheduled \citep{lee2024talr}; the cost of imposed sign switches
grows with hidden-weight magnitude \citep{laborieux2020metaplasticity}; gradient-direction instability depends on the
distance to the threshold \citep{liu2023ofq}; ternary LLMs suffer deadzone trapping \citep{huang2025tequila}; and a
crossing law with gradient-aligned terms and a bf16 starting lattice has been described \citep{xu2026shadow}. TerMeZO
fine-tunes BitNet models through the latents closest to the decision boundaries \citep{sifaou2026termezo}, where we find
the least gradient-enriched crossings---a contrast in design premise, not a measured contradiction. Our distinction is
the controlled measurement: realized crossers against same-level pools, separately scored gradients, and a
counterbalanced intervention on starting distance with the starting computation held fixed.

\textbf{Further related work} (\S\ref{sec:related}). In-Cell Learning keeps updates inside quantization cells so that
re-quantization reproduces released codes \citep{liu2026incell}, and adaptive rounding
\citep{nagel2020adaround,lee2023flexround} and weight preconditioning \citep{zhang2024magr} adjust weights before
quantization to minimise a loss; our projection instead adjusts bf16 inputs until unchanged exporters reproduce given codes.
Training without full-precision masters \citep{nikdan2026eco}, including stochastic-rounding updates of the low-bit
weights themselves \citep{zhao2024dqt}, removes the latents that our setting exports.

\ifarrangement
\section{The arrangement study}
\label{app:arrangement}
The arrangement study was first run outside the packaged pipeline and then re-executed under it after its original
results were known; the re-execution is a procedural check, not a blind test, and its records are the result of
record.

\textbf{Question.} Does the assignment of the released latent values affect 300-step GSM8K adaptation of
Falcon-E-1B-Base under a fixed recipe, when the initial executed weights are matched? The start, TRUE, is the released
latents executed by the training quantizer, the training start of the Falcon-E observational replication
(Appendix~\ref{app:selection}). That study's frozen verdict concerns its pre-registered analysis; the arrangement study
reuses only its training run, recipe, data and seed. TRUE is the online model, not the deployed packed model: fp32 codes
of these latents differ from the deployed codes on 28,132,333 coordinates (1.774\%, Table~\ref{tab:census}), each a
$\pm1$ that the deployed export removes.

\textbf{Arms.} Each arm is built in memory from TRUE's latents and executes bitwise-identically to TRUE at step 0 (codes,
bf16 effective weights up to the sign of a zero, per-row logits and cross-entropy, every gradient), with the latent
values placed differently.
MIRROR negates every zero-code latent: it keeps every coordinate's $|w|$ but not the signed latent distribution. PERM
permutes the values within each (row, code) group, with a fresh permutation per seed: it keeps each group's signed
multiset of latent values. DELTA moves every latent by one fp32 ulp in a random direction, the smallest meaningless
change, and serves as a sensitivity reading. PERM and DELTA leave latents within $10^{-5}$ (relative) of the threshold
untouched. A step-zero witness checks the bitwise match on the first 16 rows of each seed's batch order and also hashes
every parameter that is not a BitLinear weight; it is not evidence of equal fp32 amplitudes, since rearranging the
latents changes the summation order of the fp32 mean that sets the scale, and DELTA changes every latent.

\textbf{Recipe and seeds.} Full quantization-aware training with fp32 latents, bf16 autocast, gradient checkpointing,
PagedAdamW8bit (betas 0.9 and 0.95, no weight decay) and pooled response-only cross-entropy, at learning rate
$3.125\times10^{-6}$ for 300 steps (warm-up 30, cosine schedule, accumulation 16, clipping 1.0), with dev cross-entropy
on the 512 dev rows at steps 0, 150 and 300: the Falcon-E observational replication's recipe and data. Seeds
2026091700 (that study's), 2026101511 and 2026101512 set the data order and the PERM and DELTA draws, for twelve runs.
TRUE at the first seed must reproduce that earlier run's 300 step losses and three dev cross-entropies exactly; this
checks the trainer and is not a result.

\textbf{Stage B, the frozen training rule.} The primary outcome is the dev pooled cross-entropy at step 300, paired within a
seed: $\Delta_{X,s} = \mathrm{CE}_{X,s} - \mathrm{CE}_{\mathrm{TRUE},s}$ for $X$ in MIRROR and PERM. The DELTA
sensitivity is $\nu = \max_s |\mathrm{CE}_{\mathrm{DELTA},s} - \mathrm{CE}_{\mathrm{TRUE},s}|$, and the margin is
$m = \max(0.005~\text{nats}, 2\nu)$: the formula was frozen before training, and its value is realised from the DELTA
runs. Per arm, ``effect'' means that all three $\Delta_{X,s}$ share a sign and the smallest $|\Delta_{X,s}|$ exceeds
$m$; ``equivalent'' means that all three paired differences lie inside the realised margin $m$, on three seeds under
this recipe, a practical criterion and not statistical equivalence, latent irrelevance or code-only equivalence; and
``inconclusive'' is anything else. The same rule at step 150 is a secondary result.

\textbf{Stage A, the frozen rule before training.} On CPU, a zero-code latent's sign is compared with the descent direction
given by saved base-gradient scores, against an exact within-level sign-permutation null, in the primary cell (zero
code, under 0.5 ULP from the threshold, hidden codes excluded), pooled over tensors with a 2,000-draw tensor-cluster
bootstrap: for 2B-4T with GSM8K calibration scores and with C4 scores, and for Falcon-E-1B-Base with the observational
replication's GSM8K scores. ``Information present'' means an excess of at least 0.010 in magnitude with a 95\% interval
that excludes 0; ``equivalent to the null'' means that the interval lies inside $\pm0.005$; ``inconclusive'' is
anything else. This is not the same-level lift of \S\ref{sec:selection}, and neither statistic tests the other.

\textbf{Results.} TRUE at seed 2026091700 reproduced the Falcon-E observational replication's run exactly (all 300
logged step losses and the dev cross-entropies 1.007698, 0.390273 and 0.378208 at steps 0, 150 and 300); this is a
check of the trainer, not a result. The guard band caught no coordinate: every PERM and DELTA build at every
seed records zero latents within $10^{-5}$ (relative) of the threshold. At step 300 (primary) the observed DELTA
sensitivity is $\nu = 0.000208$ (per seed 0.000208, 0.000099, 0.000178), $2\nu = 0.000417$, and the realised margin is
$m = 0.0050$ nats, set by the 0.005 floor term; the paired differences $\mathrm{CE}_X - \mathrm{CE}_{\mathrm{TRUE}}$
are $+0.001547$, $+0.001690$ and $+0.001412$ nats for MIRROR and $+0.001196$, $+0.000666$ and $+0.001030$ nats for PERM
(seeds 2026091700, 2026101511, 2026101512), and DELTA's are $-0.000208$, $-0.000099$ and $-0.000178$. MIRROR and PERM are each ``equivalent'' relative to TRUE under the frozen rule, meaning that all three paired development cross-entropy differences lie inside the realised margin $m$ on three seeds under this recipe, a practical criterion, not statistical equivalence, latent irrelevance or code-only equivalence. At step 150 (secondary), $\nu = 0.000276$ (per seed 0.000276, 0.000228, 0.000269), $2\nu = 0.000552$ and $m = 0.0050$ nats
(the floor term); the paired differences are $+0.002220$, $+0.002357$ and $+0.001808$ nats for MIRROR, $+0.002088$,
$+0.001744$ and $+0.001392$ nats for PERM, and $+0.000276$, $-0.000228$ and $-0.000269$ for DELTA; MIRROR and PERM are again ``equivalent'' under the same practical criterion, not statistical equivalence. Stage A: for 2B-4T with
the GSM8K calibration scores (128 rows) the zero-cell excess agreement is $-0.00014$ over $N$ = 42,027,471 coordinates
(agreement 0.49986 against a null share of 0.50000; 95\% tensor-bootstrap interval [$-0.00032$, $+0.00004$]; 43.3\% of
the 104 tensors with at least 200 coordinates positive); with the C4 scores (128 rows, computed at batch 1) it is
$+0.00090$ over $N$ = 42,027,496 (agreement 0.50090; [$+0.00070$, $+0.00109$]; 76.0\% of 104 tensors positive); for Falcon-E-1B-Base
with the observational replication's GSM8K scores it is $+0.00029$ over $N$ = 29,331,365 (agreement 0.50029; [$+0.00012$, $+0.00047$];
61.4\% of 101 tensors positive). All three are ``equivalent to the null'' under the frozen rule, meaning that the 95\% tensor-bootstrap interval lies inside $\pm0.005$, a practical margin that does not establish latent-phase homogeneity, the absence of subgroup effects or code-only adaptation equivalence; the C4 and Falcon-E intervals exclude 0.

\textbf{Descriptive readouts.} Descriptive, post hoc in the original study and declared before the re-execution;
reported beside the frozen verdicts, never instead of them. MIRROR and PERM share TRUE within each seed and steps 150
and 300 share trajectories: the 12 treatment differences are not 12 independent replications; the row-bootstrap
intervals resample development rows and are conditional on the fitted runs; ratios to $\nu$ are not effect sizes or tests; Jaccard $|T\cap X|/|T\cup X|$ and the share of TRUE's set $|T\cap X|/|T|$ are different denominators. All 12
treatment differences are positive (TRUE lower); the mean step-300 differences are $+0.001549$ nats for MIRROR,
$+0.000964$ for PERM and $-0.000162$ for DELTA; the step-300 ratios to $\nu$ are 6.8--8.1 for MIRROR and 3.2--5.7 for
PERM. Per-seed paired row-bootstrap 95\% intervals at step 300: MIRROR [$+0.0007$, $+0.0024$], [$+0.0009$, $+0.0025$],
[$+0.0006$, $+0.0022$]; PERM [$+0.0003$, $+0.0021$], [$-0.0002$, $+0.0015$], [$+0.0002$, $+0.0019$]; DELTA's all
include 0. Training losses first differ from TRUE's at logged step index 4 (zero-based) in all nine non-TRUE runs; the
mean gap over the last 100 steps is $+0.0011$ to $+0.0019$ for MIRROR, $+0.0008$ to $+0.0015$ for PERM and $-0.0002$
to $-0.0001$ for DELTA. Changed coordinates at step 300 (seeds in order): TRUE 5,955,196 / 5,968,601 / 5,968,456;
MIRROR 5,919,806 / 5,932,718 / 5,932,665; PERM 5,924,315 / 5,938,473 / 5,936,990; DELTA 5,955,511 / 5,972,855 /
5,971,078 (additions and removals by type at steps 150 and 300 from \nolinkurl{analysis.json}'s \code{code\_changes});
there are no net sign flips relative to initialization at the measured checkpoints (every logged count is 0; nothing
is known between checkpoints). Step-300 overlaps with TRUE's changed set, as Jaccard $|T\cap X|/|T\cup X|$ beside the share of TRUE's changed set $|T\cap X|/|T|$: PERM 0.0233 / 0.0232 / 0.0232 and 4.55\% / 4.53\% / 4.53\%; MIRROR
0.1565 / 0.1571 / 0.1559 and 26.98\% / 27.07\% / 26.90\%; DELTA 0.8098 / 0.8277 / 0.8121 and 89.49\% / 90.61\% /
89.65\%; TRUE at the first seed against the second and third seeds 0.3680 / 0.3708 and 53.86\% / 54.16\%.

\textbf{Reading.} Under this Falcon-E-1B/GSM8K/300-step recipe, MIRROR and PERM produce consistently higher observed
dev CE than TRUE, while all paired differences remain inside the frozen practical margin. The interventions
substantially alter which coordinates differ from initialization at the endpoint. This neither establishes a
margin-exceeding benefit of the released assignment nor supports latent irrelevance, statistical equivalence,
code-only equivalence, or generalization to other recipes/tasks.

\textbf{Concordance.} The packaged re-execution equalled the original, unpackaged run's records in every specified
compared field and array (every training record, all 24 sparse code-change sets coordinate for coordinate, every
qualification record, every field of the frozen Stage B analysis and every shared field of each Stage A result; the
volatile timing fields, the score path and the declared witness-extension keys excluded): equality of the compared
records under the same code, data, seeds and hardware, not raw byte identity of every file and not an independent
replication. The verdicts and numbers above are the re-execution's, the result of record.

\textbf{Cost.} 13,331.8 GPU-seconds (3.70 GPU-hours) for the run and 107.7 GPU-seconds for the two build checks, on one
RTX 4090; Stage A ran on CPU (68 s).

\textbf{Limitations.} One model, one task, one learning rate, 300 steps, three seeds and dev cross-entropy only; no
accuracy was measured, there is no code-only arm and no frozen-code control, and the margins are practical criteria,
not the result of a power analysis. $\nu$ comes from three DELTA runs and is a sensitivity reading, not an uncertainty
estimate. Within a seed the treatments share TRUE and the two checkpoints share trajectories, so the paired differences
are not independent replications. The released latents are bf16-rounded, so their sub-ULP phase is lost in the release
itself. The re-run re-executes a study whose results were known; it is a procedural check, not an independent
replication.
\fi

\section{The exports in bitnet.cpp 01eb415 (CPU)}
\label{app:e34d}
We converted the documented and projected exports of 2B-4T's C1 and Falcon-E's A1 endpoints (and Falcon-E's fixed
export) with bitnet.cpp's documented GGUF routes (preprocess, convert, \code{llama-quantize I2\_S}; the Falcon-E inputs
normalized by the release tokenizer files and a top-level \code{rope\_theta}, since the unmodified route fails at
tokenizer loading on the labs' tool output) and generated the 512 dev rows one unpadded row at a time with
\code{llama-server} (greedy, 16 threads, token ids taken from the server's debug log and cross-checked against the
returned pieces). The runtime was screened on the released bases against an HF fp32 reference, a precision selected
after seeing the released-base comparisons (V1 5/8 and 7/8 prompts; V2 top-1 290/302 and 296/303; bf16 rows 3/8,
289/302 and 5/8, 299/303), and a decode-path diagnostic agreed at 245/248 and 246/248 positions; passing this screen is
evidence on those inputs, not a runtime-equivalence guarantee. Code census: the documented route kept the base code at
12,320,959 of 12,905,600 (2B-4T) and 3,987,515 of 5,955,167 (Falcon-E) of fine-tuning's code changes, the same codes at
every coordinate as the HF-reload and onebitllms export paths; the projected exports carry the online endpoint codes at
all 2,084,044,800 and 1,585,446,912 coordinates (targets \code{6ef86536…} and \code{3231f2d4…}), with amplitudes equal to
each input's declared representation (fp32(1/s), fp32(1/weight\_scale)) that differ from the online endpoint's fp32
amplitude by at most $2.1\times10^{-5}$ (2B-4T) and $5.5\times10^{-3}$ (Falcon-E) relative; code equality is not output
equality. The documented states' amplitudes are likewise as declared and differ from the online endpoint's fp32
amplitude by at most $1.6\times10^{-5}$ (2B-4T) and $5.5\times10^{-3}$ (Falcon-E) relative. Converted instead from
2B-4T's fp32 masters (a converter-input experiment), the runtime's codes differ from the GPU-captured endpoint codes at
920 of 2,084,044,800 coordinates, all within $2u$ of the rounding threshold ($u = 2^{-24}$; 542 exact ties, 378 within
$u$), under the route's preprocess (torch 2.2.2+cpu, 16 threads); this is a different stage and environment from the
852-coordinate figure of \S\ref{sec:remedies} and is not merged with it. Accuracy (Table~\ref{tab:e34d}): the
documented export scored below the projected export in bitnet.cpp 01eb415 (CPU) under the strict rule for both models
(2B-4T's last-number accuracy is higher under the documented export, as in PyTorch); Falcon-E's fixed and projected
exports show no detectable difference. Against the bf16-autocast PyTorch online evaluation (batch 64, left-padded, GPU;
a different implementation, batch and device) the projected exports in bitnet.cpp show no detectable difference and are
non-inferior: $\delta = 4$ pp is the inherited resolution-based margin derived from these 512 dev rows' export contrasts
(median 95\% half-width, rounded down); it is used for continuity, not re-derived. Marginal 95\% paired row bootstrap
intervals (20,000 draws, one index matrix shared by all contrasts, so they are correlated); no detectable difference is
not equivalence. Determinism repeats of 32 rows per state reproduced every token; timings are descriptive, measured on
a shared host without an exclusive reservation (load average 14--19 during generation). Two diagnostics are counted
separately: ambiguous returned pieces, 0 and 0 (2B-4T documented, projected) and 18, 18 and 25 (Falcon-E documented,
projected, fixed), and omitted probability positions, 12 and 0 (2B-4T) and 18, 18 and 25 (Falcon-E), each decided by
the logged token ID; the decode-path diagnostic is a replay of inputs seen in the build checks, not a fresh blind test.
One technical revision: an external HTTP \code{HEAD /} probe reached the server's loopback port
during row 222 of Falcon-E's projected-export generation, the adapter refused the attempt, and the phase was
regenerated in full without code change (the refused attempt is retained, never scored). The result review was a
substitute independent review; the designated reviewer's confirmation was obtained on 2026-10-23. Separately, a static
reading of the source of Microsoft's GPU W2A8 decode kernel suggests a q/k/v scale-band assignment issue for 2B-4T's
grouped-query shapes; this static source observation is unmeasured, and we defer it to future work.
\begin{table}[H]
\centering\scriptsize
\setlength{\tabcolsep}{3pt}
\caption{The exports executed in bitnet.cpp 01eb415 (CPU) on the 512 dev rows: differences $x - y$ in points, with marginal 95\% paired row-bootstrap intervals (20,000 draws; one index matrix shared by all contrasts, so they are correlated). Strict accuracies: 2B-4T documented 30.27\%, projected 57.81\% (online 58.40\%); Falcon-E documented 0.78\%, projected 60.16\%, fixed 58.79\% (online 58.79\%). Answer-format rates: 40.8\% and 98.0\% (2B-4T documented, projected); 21.3\%, 97.3\% and 97.7\% (Falcon-E documented, projected, fixed). The Falcon-E inputs were normalized. The last two rows are cross-implementation comparisons (bf16-autocast PyTorch on GPU, batch 64 left-padded, against single unpadded rows on CPU); $\delta$ = 4 pp is the inherited resolution-based margin derived from these 512 dev rows' export contrasts (median 95\% half-width, rounded down); it is used for continuity, not re-derived. No detectable difference is not equivalence. All timings were measured on a shared host and are descriptive. The plain-language results map to the registry's labels in Table~\ref{tab:key}.}
\label{tab:e34d}
\begin{adjustbox}{max width=\textwidth}
\begin{tabular}{llll}
\toprule
contrast ($x - y$, 512 dev rows) & $\Delta$ strict & $\Delta$ last-number & result \\
\midrule
2B-4T, documented $-$ projected & \iv{$-27.54$}{$-32.81$, $-22.07$} & \iv{$+5.27$}{$+0.39$, $+10.16$} & documented below projected \\
Falcon-E, documented $-$ projected & \iv{$-59.38$}{$-63.67$, $-55.08$} & \iv{$-14.45$}{$-19.73$, $-8.98$} & documented below projected \\
Falcon-E, documented $-$ fixed (secondary) & \iv{$-58.01$}{$-62.30$, $-53.71$} & \iv{$-13.09$}{$-18.36$, $-7.81$} & lower \\
Falcon-E, fixed $-$ projected (secondary) & \iv{$-1.37$}{$-3.91$, $+1.17$} & \iv{$-1.37$}{$-3.91$, $+1.17$} & no detectable difference \\
2B-4T, projected in bitnet.cpp $-$ online PyTorch & \iv{$-0.59$}{$-2.93$, $+1.76$} & \iv{$-0.59$}{$-2.93$, $+1.76$} & no detectable difference; non-inferior at $\delta = 4$ pp \\
Falcon-E, projected in bitnet.cpp $-$ online PyTorch & \iv{$+1.37$}{$-0.98$, $+3.71$} & \iv{$+1.37$}{$-0.98$, $+3.71$} & no detectable difference; non-inferior at $\delta = 4$ pp \\
\bottomrule
\end{tabular}
\end{adjustbox}
\end{table}

\end{document}